\documentclass[onecolumn,english]{IEEEtran}
\usepackage[T1]{fontenc}
\usepackage[latin9]{inputenc}
\usepackage{units}
\usepackage{amsthm}
\usepackage{amsmath}
\usepackage{amssymb}
\usepackage{graphicx}
\usepackage{booktabs}
\usepackage[noend]{algpseudocode} 
\usepackage{algorithm}

\usepackage[usenames,dvipsnames,svgnames]{xcolor} 

\usepackage[caption=false,font=footnotesize]{subfig}
\usepackage{url}
\usepackage{pifont}

\usepackage{multirow}

\newif\ifIEEEsubmission
\IEEEsubmissionfalse

\newlength{\figWidth}
\ifIEEEsubmission
\else
\fi

\usepackage[normalem]{ulem} 

\makeatletter
\theoremstyle{plain}
\newtheorem{thm}{\protect\theoremname} 
\theoremstyle{plain}
\newtheorem{lem}[thm]{\protect\lemmaname}
\newtheorem{cor}[thm]{\protect\corollaryname}

\newtheorem*{remark}{Remark}

\usepackage{amsmath}

\DeclareMathOperator*{\argmin}{arg\,min}
\DeclareMathOperator{\diag}{diag}
\DeclareMathOperator{\trace}{tr}

\newcommand{\order}{\mathcal{O}}              

\newcommand{\x}{\mathbf{x}}
\renewcommand{\xi}{{\x}_{i}}
\newcommand{\X}{\mathbf{X}}
\newcommand{\Y}{\mathbf{Y}}
\newcommand{\y}{\mathbf{y}}
\newcommand{\w}{\mathbf{w}}
\newcommand{\R}{\mathbb{R}}    
\newcommand{\RR}{\mathbf{R}}   
\renewcommand{\P}{\mathbb{P}}  
\newcommand{\E}{\mathbb{E}}    
\newcommand{\Z}{\mathbf{Z}}

\newcommand{\maxRow}{\text{max-row}}
\newcommand{\eye}{\mathbf{I}}
\newcommand{\maxCol}{\text{max-col}}

\newcommand{\rand}{\mathbf{\Omega}}

\usepackage[noadjust]{cite}
\usepackage[caption=false]{subfig}

\@ifundefined{showcaptionsetup}{}{%
 \PassOptionsToPackage{caption=false}{subfig}}
\usepackage{subfig}
\makeatother

\usepackage{babel}
\providecommand{\lemmaname}{Lemma}
\providecommand{\theoremname}{Theorem}
\providecommand{\corollaryname}{Corollary}
\providecommand{\propositionname}{Proposition}

\DeclareMathAlphabet{\mathbbb}{U}{bbold}{m}{n}

\begin{document}

\title{Preconditioned Data Sparsification for Big Data with Applications to PCA and K-means}

\author{Farhad Pourkamali-Anaraki and Stephen Becker%
\thanks{Farhad Pourkamali-Anaraki is with the Department
of Electrical, Computer, and Energy Engineering, University of Colorado Boulder, Boulder, CO 80309 USA (e-mail: farhad.pourkamali@colorado.edu). 

Stephen Becker is with the Department of Applied Mathematics, University of Colorado Boulder, Boulder, CO 80309 USA (e-mail: stephen.becker@colorado.edu).
}}
\maketitle
\begin{abstract}
We analyze a compression scheme for large data sets that randomly keeps a small percentage of the components of each data sample. The benefit is that the output is a sparse matrix and therefore subsequent processing, such as PCA or K-means, is significantly faster, especially in a distributed-data setting. Furthermore, the sampling is single-pass and applicable to streaming data. The sampling mechanism is a variant of previous methods proposed in the literature combined with a randomized preconditioning to smooth the data. We provide guarantees for PCA in terms of the covariance matrix, and guarantees for K-means in terms of the error in the center estimators at a given step. We present numerical evidence to show both that our bounds are nearly tight and that our algorithms provide a real benefit when applied to standard test data sets, as well as providing certain benefits over related sampling approaches.
\end{abstract}
\section{Introduction\label{sec:Introduction}}

Principal component analysis (PCA) is a classical unsupervised analysis method commonly used in all quantitative disciplines~\cite{PCA_Book}. Given $n$ observations of $p$-dimensional data 
$\left\{ \x_{i}\right\} _{i=1}^n\in\R^p$, standard algorithms to compute PCA require $\order(p^2n)$ flops (we assume throughout the paper that $n\ge p$), which is expensive for large $p$ and/or $n$. The problem is exacerbated  in situations with high communication cost, as in distributed data settings such as Hadoop/MapReduce clusters or as in sensor networks. An extreme case is when the data rate is so large that we turn to ``streaming'' algorithms which examine each new data entry and then discard it, i.e., take at most one pass over the data~\cite{DataStreamsBook}.

Another commonly used unsupervised analytic task is  clustering, which refers to identifying groups of similar data samples in a data set. Running a simple clustering method such as the K-means algorithm turns out to be infeasible for distributed and streaming data~\cite{BigDataSigMagazine}. Therefore, a recent line of research aims at developing scalable learning tools for ``big data'' and providing fundamental insights and tradeoffs involved in these algorithms.

This paper proposes a specific mechanism to reduce the computational, storage, and communication burden in unsupervised analysis methods such as PCA and K-means clustering, which requires only one pass over the data. The first step is a pre-processing step designed to precondition the data and smooth out large entries, which is based on the well-known fast Johnson-Lindenstrauss result~\cite{FJLT}. We summarize existing results about this preconditioning and describe how it benefits our approach. The second step is element-wise random sampling: choosing to keep exactly $m$ out of $p$ entries (without replacement) per sample. This step can be viewed as forming a sampling matrix $\RR_i\in\R^{p\times m}$ which contains $m$ distinct canonical basis vectors drawn uniformly at random, and multiplying each preconditioned sample $\mathbf{x}_i$ on the left by $\RR_i\RR_i^T$ (the projection matrix onto the subspace spanned by the columns of $\RR_i$). The two steps of our approach, preconditioning and sampling, can be easily combined into single pass over the data which eliminates the need to revisit past entries of the data.

We then analyze, in a non-Bayesian setting, the performance of our estimators and provide theorems that show as $n$ grows with $p$ fixed, the bounds hold with  
high probability
\footnote{in the sense of~\cite{randomMatrix}} if
 $m=\order(\log n/n)$,
 cf. Corollary~\ref{cor:sampleMean}.
 This means that as we collect more samples, the size of the incoming data increases proportional to $n$ but the amount of data we must store only increases like $\log n$, i.e., we compress with ratio $n/\log n$. This is possible because of the careful sampling scheme --- if one were to simply keep some of the data points $\xi$, then for a small error in the $\ell_\infty$ norm, one needs to keep a \emph{constant} fraction of the data.

Similar proposals, based on dimensionality reduction or sampling, have been proposed for analyzing large data sets, e.g.,~\cite{DataBaseFriendlyRandomProjection,Martinson_SVD,RandomizedAlgorithmsforMatricesAndData}. Many of these schemes are based on multiplying the data on the left by a single random matrix $\rand$ and recording $\rand \X$ where $\X=[\x_1,\ldots,\x_n]$ is the $p\times n$ matrix with data samples as columns, $\rand$ is $m\times p$ with $m<p$, and the columns of $\rand \X$ are known as sketches or compressive measurements. The motivation behind sketching is the classical  Johnson-Lindenstrauss lemma \cite{JLThm}  which states that the scaled pairwise distances between low-dimensional sketches are preserved to within a small tolerance for some random matrices $\rand$. Moreover, the more recent field of compressive sensing  demonstrates that low-dimensional compressive measurements $\rand \mathbf{X}$ contain enough information to recover the data samples under the assumption that $\mathbf{X}$ is sparse in some known basis representation~\cite{CSCandes}.   

However, for PCA, the desired output is the eigendecomposition of $\X\X^T$ or, equivalently, the left singular vectors of $\X$, for which we cannot typically impose structural constraints. Clearly, a single random matrix $\rand^{T}\in\R^{p\times m}$ with $m<p$ only spans at most an $m$-dimensional subspace of $\R^{p}$, and after the transformation $\rand\X$, information about the components of the left singular vectors within the orthogonal complement subspace will be lost. Another technique to recover the left singular vectors is to also record $\X\rand'$ for another random matrix $\rand'$ of size $n \times m'$ with $m'<n$~\cite{Martinson_SVD}, and with careful implementation, it is possible to record $\rand\X$ and $\X\rand'$ in a one-pass algorithm. However, even in this case, it is not possible to return something as simple as a center estimator in the K-means algorithm without making a second pass over the data.

Our approach circumvents this since each data sample $\x_i$ is multiplied by its own random matrix $\RR_i\RR_i^T$, where $\RR_i\in\R^{p\times m}$ consists of a
randomly chosen subset of $m$ distinct canonical basis vectors. We show that our scheme leads to one-pass algorithms for unsupervised analysis methods such as PCA and K-means clustering, meanwhile containing enough information to recover principal components and cluster centers without imposing additional structural constraints on them. 

We expand on comparisons with relevant literature in more detail in Section~\ref{sec:related_work}, summarize important results about the preconditioning transformation and sub-sampling in Section~\ref{sec:precondition}, provide theoretical results on the sample mean estimator and covariance estimator in sections~\ref{sec:center} and~\ref{sec:covariance} respectively, then focus on results relevant for K-means clustering in Section~\ref{sec:CK-means} and finish the paper with numerical experiments in Section~\ref{sec:numerical_experiments} and a conclusion.

\subsection{Setup}\label{sec:setup}
In this paper, we consider a non-Bayesian data setting where we make no distributional assumptions on the set of data samples $\X=[\x_1,\ldots,\x_n]\in\R^{p\times n}$. For each sample $\x_{i}$, we form a sampling matrix $\RR_i\in\R^{p\times m}$, where the $m$ columns are chosen uniformly at random from the set of canonical basis vectors in $\R^{p}$ without replacement. Thus, we use $m$ for the intrinsic dimensionality of the compressed samples, and write the compression factor as $\gamma=\frac{m}{p}$.

We name column vectors by lower-case bold letters and matrices by upper-case bold letters. Our results will involve bounds on the Euclidean and maximum norm in $\R^p$, denoted $\|\x\|_2$ and $\|\x\|_\infty$ respectively, as well as the spectral and Frobenius norms on the space of $p\times n$ matrices, denoted $\|\X\|_2$ and $\|\X\|_F$ respectively. 
Recall that $\|\x\|_\infty \le \|\x\|_2 \le \sqrt{p}\|\x\|_\infty$. We use $\|\X\|_\text{max}$ to denote the maximum absolute value of the entries of a matrix, $\|\X\|_\maxRow$ to be the maximum $\ell_2$ norm of the rows of a matrix, i.e., $\|\X\|_\maxRow = \|\X\|_{2\rightarrow \infty} = \sup_{\y\neq 0} \frac{\|\X\y\|_\infty}{\|\y\|_2}$, and $\|\X\|_\maxCol=\|\X\|_{1\rightarrow 2}$ to be the maximum $\ell_2$ norm of the columns of a matrix.

Let $\mathbf{e}_i$ denote the $i$-th vector of the canonical basis in $\R^{p}$, where entries are all zero except for the $i$-th one which is $1$. In addition, $\diag(\x)$ returns a square diagonal matrix with the entries of vector $\x$ on the main diagonal, and $\diag(\X)$ denotes the matrix formed by zeroing all but the diagonal entries of matrix $\X$. We also represent the entry in the $i$-th row and the $j$-th column of matrix $\X$ as $X_{i,j}$.

\subsection{Contributions}\label{sec:contributions}
In this paper, we introduce an efficient data sparsification framework for large-scale data applications that does not require incoherence and distributional assumptions on the data. Our approach exploits randomized orthonormal systems to find informative low-dimensional representations in a single pass over the data. Thus, our proposed compression technique can be used in streaming applications, where data samples cannot fit into memory~\cite{MemoryLimited}. We present results on the properties of sampling matrices and randomized orthonormal systems in Theorem~\ref{thm:Properties_Sketching} and Section~\ref{sec:precondition}, respectively.

To demonstrate the effectiveness of our framework, we investigate the application of preconditioned data sparsification in two important unsupervised analysis methods, PCA and K-means clustering. Two unbiased estimators are introduced to recover the sample mean and covariance matrix from the compressed data. We provide strong theoretical guarantees on the closeness of the estimated and true sample mean in Theorem~\ref{thm:sampleMean}. Moreover, we employ recent results on exponential concentration inequalities for matrices~\cite{tropp2015introduction} to obtain an exponentially decaying bound on the probability that the estimated covariance matrix
deviates from its mean value in Theorem~\ref{thm:covariance}. Numerical simulations are provided to
validate the tightness of the concentration bounds derived in this paper. Furthermore, our theoretical results reveal the connections between accuracy and important parameters, e.g., the compression factor and the underlying structure of data.

We also examine the application of our proposed data sparsification technique in the K-means clustering problem. In this case, we first define an optimal objective function based on the Maximum-Likelihood estimation. Then, an algorithm called sparsified K-means is introduced to find assignments as well as cluster centers in one pass over the data. Theoretical guarantees on the clustering structure per iteration are provided in Theorem~\ref{thm:accuracy-sparse-k-means}. Finally, we present extensive numerical experiments to demonstrate the effectiveness of our sparsified K-means algorithm on large-scale data sets containing up to $10$ million samples.

Beyond just the two unsupervised learning examples, we advocate the general usage of preconditioning followed by sampling. Sampling methods have long been popular due to their simplicity, but unless appropriate weighted sampling distributions are used, the accuracy of these methods is low. The use of the preconditioning obviates the need for weighted distributions, and is easy to analyze and implement.
\section{Related Work}\label{sec:related_work}
Our work has a close connection to the recent line of research on signal processing and information retrieval from compressive measurements that assumes one only has access to low-dimensional random projections of the data~\cite{DavenportCSSignal,SketchedSVD,park2014modal}. For example, Eldar and Gleichman~\cite{BlindCS_2010} studied the problem of learning basis representations (dictionary learning) for the data samples $\X\in\R^{p\times n}$ under the assumption that a single random matrix $\RR\in\R^{p\times m}$, $m<p$, is used for all the samples. It was shown that $\RR^{T}\X$ does not contain enough information to recover the basis representation for the original data samples, unless structural constraints such as sparsity over the set of admissible representations are imposed. This mainly follows from the fact that $\RR$ has a non-trivial null space and, without imposing constraints, we cannot recover the information about the whole space $\R^{p}$. For example,  consider a simple case of a rank-one data matrix, $\X=\sigma\mathbf{u}\mathbf{v}^T$, where the goal is to recover the single principal component $\mathbf{u}\in\R^p$ from $\RR^{T}\X$. The singular value decomposition of $\RR^{T}\X$ results in $\RR^{T}\X=\sigma\mathbf{u}'\mathbf{v}^T$, where $\mathbf{u}'=\RR^{T}\mathbf{u}\in\R^m$, so we have retained information on  $\mathbf{v}$. However, accurate estimation of $\mathbf{u}$ from $\mathbf{u}'=\RR^{T}\mathbf{u}$ is not possible, unless additional constraints are imposed on $\mathbf{u}$. The line of work~\cite{StuderDL,Pourkamali_CKSVD,Pourkamali_DL_SampTA,NewGuaranteesBCS} addressed this issue, in the dictionary learning setting, by observing each data sample $\x_i$ through \emph{distinct} random matrices.

Another line of work considers covariance estimation and recovery of principal components (PCs) based on compressive measurements~\cite{Fowler,InvarianceOfPCs,Pourkamali_ICASSP_2014,CovEstGoldsmith,Pourkamali_ICML_2014,CovEstimationCSAarti}. In particular, Qi and Hughes~\cite{InvarianceOfPCs} proposed a method for recovering the mean and principal components of $\X=[\x_1,\ldots,\x_n]$ from compressive measurements $\RR_{i}^{T}\x_i\in\R^{m}$, $i=1,\ldots,n$, where $\RR_i\in\R^{p\times m}$ with $m<p$ is a random matrix with entries drawn i.i.d.~from the Gaussian distribution. This method requires computing the projection matrix onto the subspace spanned by the columns of $\RR_i$ which is $\mathbf{P}_i=\RR_i(\RR_i^{T}\RR_i)^{-1}\RR_i^{T}$. Then, each $\mathbf{P}_i\x_i$ can be directly computed from the compressive measurements since $\mathbf{P}_i\x_i=\RR_i(\RR_i^{T}\RR_i)^{-1}\RR_i^{T}\x_i$. It has been shown that the mean and principal components of $\mathbf{P}_i\x_i\in\R^{p}$, $i=1,\ldots,n$, converge to the mean and principal components of the original data (up to a known scaling factor) as the number of data samples $n$ goes to the infinity. However, this method is computationally inefficient because 
even if each (pseudo-)random matrix $\RR_i$, $i=1,\ldots,n$, is implicitly stored by a seed, the computational cost of the matrix multiplies and inversions is high.

Their work was extended in~\cite{Pourkamali_ICML_2014}, where the authors studied the problem of performing PCA on $\RR_i\RR_i^{T}\x_i$ (eliminating the matrix inversion) for a general class of random matrices, where the entries of $\RR_i$ are drawn i.i.d.~from a zero-mean distribution with finite moments. Two estimators were proposed to estimate the mean and principal components of the original data with statistical guarantees for a finite number of samples $n$. Moreover, it was shown that one can use very sparse random matrices~\cite{VerySparseRandom} to increase the efficiency in large-scale data sets. However, generating $n$ unstructured random matrices, with $p\times m$ entries in each, can still be memory/computation inefficient for high-dimensional data sets with $p$ large. Another disadvantage of the prior work~\cite{InvarianceOfPCs} and~\cite{Pourkamali_ICML_2014} is that the result only holds for data samples drawn from a specific probabilistic generative model known as the spiked covariance model, which may not be informative for real-world data sets. 

In the fields of theoretical computer science and numerical linear algebra, there are several alternative lines of work that are related to our proposed approach and we discuss them below in detail.

\subsection{Comparison with column sampling approaches}\label{sec:ColumnSampling}
\newcommand{\UU}{\mathbf{U}}
The most natural compression scheme for large-scale data sets would be to directly select a small subset of the original data and then perform data analytic tasks on the reduced subset. The two natural distributions for column sampling are the uniform distribution and a non-uniform data-dependent distribution based on so-called statistical leverage scores~\cite{RandomizedAlgorithmsforMatricesAndData}\footnote{Sampling according to column norm is a common third option, but is generally inferior to leverage-score based sampling}. The former method, uniform sampling, is a simple one-pass algorithm but it will perform poorly on many problems if there is any structural non-uniformity in the data~\cite{StatPersLevMahoney}. 

To see this, we recreate the numerical experiment of~\cite{StatPersLevMahoney} and compare the accuracy of left singular vectors/principal components (PCs) estimated using our proposed approach with those estimated after uniform column sampling. We set the parameters $p=512$ and $n=1024$ and consider $1000$ runs for different values of the compression factor $\gamma=\frac{m}{p}$. In each run, we generate a data matrix $\X\in\R^{p\times n}$ from the multivariate $t$-distribution with $1$ degree of freedom and covariance matrix $\mathbf{C}$ where $C_{ij}=2\times 0.5^{|i-j|}$. In our approach, we precondition the data as described in \S\ref{sec:precondition} and then keep exactly $m$ out of $p$ entries for each data sample to obtain a sparse matrix. To have a fair comparison, we consider randomly selecting $2m$ columns of $\X$ because $n/p=2$  and our sparse matrix has exactly $2mp$ nonzero entries. We measure the accuracy of the estimated PCs based on the explained variance~\cite{MemoryLimited}: given estimates of $k$ PCs $\widehat{\UU}\in\R^{p\times k}$ (we take $k=10$), the fraction of explained variance is defined as $\trace(\widehat{\UU}^{T}\X\X^{T}\widehat{\UU})/\trace(\X\X^{T})$, and closeness of this fraction to $1$ represents higher accuracy. 

\begin{figure}
 	\centering
 	\includegraphics[width=\figWidth]{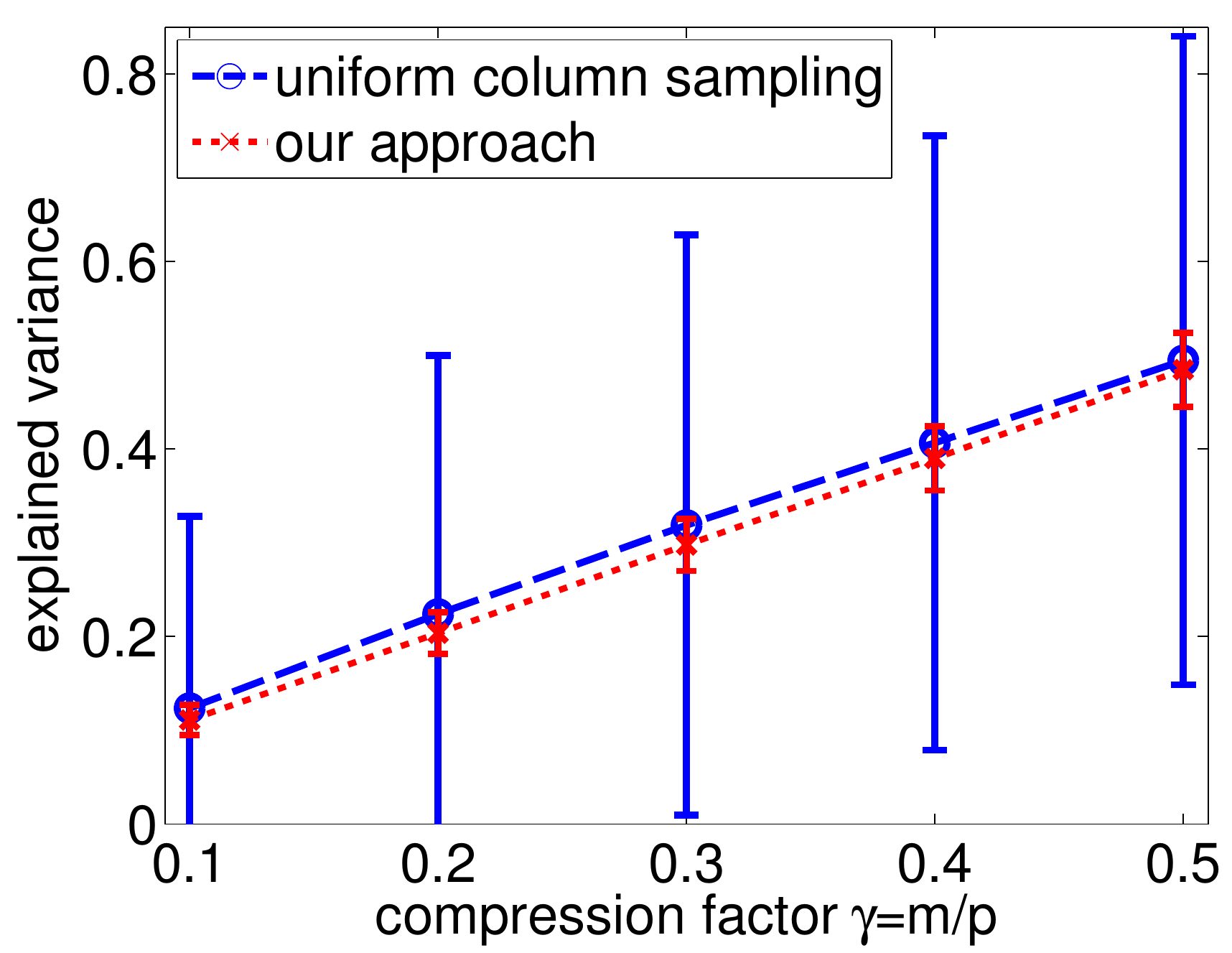}
 	\caption{Accuracy of estimated PCs via one-pass methods: uniform column sampling and our proposed precondition+sparsification approach. We plot the mean and standard deviation of the explained variance over $1000$ runs for each value of $\gamma$. The standard deviation for our approach is significantly smaller 
          compared to the uniform column sampling. 
 		\label{fig:column-sampling}
 	}
 \end{figure}

Fig.~\ref{fig:column-sampling} reports the mean and standard deviation of the explained variance over $1000$ trials. For both approaches, the average explained variance approaches $1$ as $\gamma\rightarrow 1$, as expected, and uniform column sampling is slightly more accurate than our approach.
However, uniform column sampling has an extremely high variance for all values of $\gamma$, e.g., the standard deviations for $\gamma=0.1$, $0.2$, $0.3$ are $0.20$, $0.28$, and $0.31$ respectively.
On the other hand, the standard deviation for our approach is significantly smaller for all values of $\gamma$ (less than $0.04$),
 due to the preconditioning.
Thus the \emph{worst-case} performance of our approach is reasonable, while the worst-case performance of column sampling may be catastrophically bad.

 The second common method for column sampling is based on a data-dependent non-uniform distribution and has received great attention in the development of randomized algorithms such as low-rank matrix approximation. These data-dependent sampling techniques are typically computationally expensive and a SVD on the data matrix is required. Recently, there are variants that can be computed more efficiently such as~\cite{FastApproxCohLev}. However, these algorithms compute the sampling distribution in at least one pass of all the samples, and a second pass is required to actually sample the columns based on this distribution. Thus at least two passes are required and these algorithms are not suitable for  streaming~\cite{Ghashami_Det_low_rank}.
\subsection{Comparison with other element-wise sampling approaches}
Analysis of sparsification of matrices for the purpose of fast computation of low-rank approximations goes back to at least Achlioptas and McSherry~\cite{AchlioptasMcSherry2001,AchlioptasMcSherry}, who propose independently keeping each entry of the matrix in either a uniform fashion or a non-uniform fashion. In the former, each entry of the data matrix is kept with the same probability, whereas in the non-uniform case, each entry is kept with probability proportional to its magnitude squared. Under the latter scheme, 
the expected number of nonzero entries can be bounded but 
one does not have precise control on the exact number of nonzero entries.

Recently, Achlioptas et al.~\cite{NearOptimalSampling} have considered a variant of element-wise sampling where a weight is assigned to each entry of the data matrix according to its absolute value normalized by the $\ell_{1}$ norm of the corresponding row, and then a fixed number of entries is sampled (with replacement) from the matrix. Computing the exact $\ell_{1}$ norms of the rows requires one pass over the data and consequently this method requires two passes over the data. It is shown empirically  that disregarding the normalization factor performs quite well in practice, thus yielding a one-pass algorithm. Calculating guarantees on their one-pass variant is an interesting open-problem.
 
 \subsection{K-means clustering for big data}
Clustering is a commonly used unsupervised learning task that reveals the underlying structure of a data set by splitting the data into groups, or clusters, of similar samples. It has applications ranging from search engines and social network analysis to medical and e-commerce domains. Among clustering algorithms, K-means~\cite{Bishop} is one of the most popular clustering algorithms~\cite{TopKMeans}. K-means is an iterative expectation-maximization type algorithm in which each cluster is associated with a representative vector or cluster center, which is known to be the sample mean of the vectors in that cluster. In each iteration of K-means, data samples are assigned to the nearest cluster centers (usually based on the Euclidean norm) and cluster centers are then updated based on the most recent assignment of the data. Therefore, the goal of K-means is to find a set of cluster centers as well as assignment of the data.  
 
Despite the simplicity of K-means, there are several challenges in dealing with modern large-scale data sets. The high dimensionality and large volumes of data make it infeasible to store the full data in a centralized location or communicate the data in distributed settings~\cite{BigDataSigMagazine} and these voluminous data sets make K-means computationally inefficient~\cite{Randomized_Dim_K_means}.

Recent approaches to K-means clustering for big data have focused on selecting the most informative features of the data set and performing K-means on the reduced set. For example, Feldman et al.~\cite{K-means-Feldman} introduced coresets for K-means, in which one reduces the dimension of the data set from $p$ to $m$ by projecting the data on the top $m$ left singular vectors (principal components) of $\X$, although this requires the SVD of the data matrix and so does not easily apply to streaming scenarios. To reduce the cost of computing exact SVD, the authors in~\cite{cohen2015dimensionality} proposed using approximate SVD algorithms instead.

Other prior works, such as \cite{Randomized_Dim_K_means,SketchAndValidateKMeans}, have used randomized schemes for designing efficient clustering algorithms. In particular, Mahoney et al.~\cite{Randomized_Dim_K_means} proposed two provably accurate algorithms to reduce the dimension of the data by selecting a small subset of $m$ rows of the data matrix $\X\in\R^{p\times n}$ (feature selection) or multiplying the data matrix from the left by a random matrix $\rand\in\R^{m\times p}$ (feature extraction). Afterwards, the K-means algorithm is applied on these $n$ data samples in $\R^{m}$ to find the assignment of the original data. In feature selection, the rows of $\X$ are sampled via a non-uniform distribution, which requires two passes over the data. Then, having the distribution, it requires one more pass to actually sample $m$ rows which leads to a three-pass algorithm. 

Furthermore, in both feature selection and feature extraction algorithms, K-means is applied on the projected data in $\R^{m}$ so
there is no ready estimate of the cluster centers in the original space $\R^{p}$. One could transform the cluster centers to $\R^p$ with the pseudo-inverse of $\rand$ (recall $\rand$ is low-rank) but this estimate is poor; see Figs.~\ref{fig:extract_1} and \ref{fig:selection_3} in \S\ref{sec:numerical_experiments}.
A better approach to calculate the cluster centers is to use the calculated assignment of vectors in the original space, which requires one additional pass over the data, and consequently neither feature selection nor feature extraction is streaming.
 
In this paper, we address the need to account for the computational/storage burden and we propose a randomized algorithm for K-means clustering, called sparsified K-means, that returns both the cluster centers and the assignment of the data in single pass over the data. To do this, we take a different approach where a randomized unitary transformation is first applied to the data matrix and we then choose $m$ out of $p$ entries of each preconditioned data uniformly at random to obtain a sparse matrix. 
Afterwards, the K-means algorithm is applied on the resulting sparse matrix and, consequently, speedup in processing time and savings in memory are achieved based on the value of compression factor $\gamma=m/p<1$.

We provide theoretical guarantees on the clustering structure of our sparsified K-means algorithm compared to K-means on the original data in each iteration. In fact, this is another advantage of our method over the previous work~\cite{Randomized_Dim_K_means}, where guarantees are available only for the final value of the~\emph{objective function} and, thus, it is not possible to directly compare the clustering structure of feature selection and feature extraction algorithms with K-means on the original data.

\section{Preliminaries}\label{sec:precondition}
\newcommand{\Hadamard}{\mathbf{H}}
\newcommand{\Diag}{\mathbf{D}}
The randomized orthonormal system (ROS) is a powerful randomization tool that has found uses in fields from machine learning~\cite{FJLT} to numerical linear algebra~\cite{Martinson_SVD} and compressive sensing~\cite{FastAndEfficientCS}. 
We use the ROS to efficiently precondition the data and smooth out large entries in the 
matrix $\X$ before sampling. It is straightforward to combine this preconditioning and sampling operation into a single pass on the data. Furthermore, we can unmix the preconditioned data because the preconditioning transformation is unitary, so by applying its adjoint we undo its effect. Because the operator is unitary, it does not affect our estimates in the Euclidean norm (for vectors) or the spectral and Frobenius norms (for matrices). Moreover, this transformation is stored implicitly and based off fast transforms, so applying to a length $p$ vector takes $\order(p\log(p))$ complexity, and applying it to a matrix is embarrassingly parallel across the columns~\cite{FastFood,sindhwani2015structured}.

Specifically, the ROS preconditioning transformation
uses matrices $\Hadamard$ and $\Diag$ and is defined 
\begin{equation}
\x\mapsto\y=\Hadamard\Diag\x\label{eq:ROS}
\end{equation}
where $\Hadamard\in\R^{p \times p}$ is a deterministic orthonormal matrix such as a Hadamard, Fourier, or DCT matrix for which matrix-vector multiplication can be implemented efficiently in $\order(p\log(p))$ complexity without the need to store the matrices explicitly. The matrix $\Diag\in\R^{p \times p}$ is a stochastic diagonal matrix whose entries on the main diagonal are random variables drawn uniformly from $\{\pm 1\}$. The matrix product $\Hadamard\Diag\in\R^{p \times p}$ is an orthonormal matrix and this mapping ensures that, with high probability, the magnitude of any entry of $\Hadamard\Diag\x$ is about $\order(1/\sqrt p)$ for any unit vector $\x$ (cf.~Thm.~\ref{thm:FJLT}). 

We now present relevant results about the preconditioning transformation $\Hadamard\Diag
$ \eqref{eq:ROS} that will be used in the next sections; more sophisticated results are in \cite{ImprovedAnalysis} and Appendix~\ref{sec:pairwiseDistance}.

\begin{thm}[Single element bound for ROS~\eqref{eq:ROS}] \label{thm:FJLT}
	Let $\x\in\R^p$, and $\y=\Hadamard\Diag\x$ a random variable from the ROS applied to $\x$, then for every $j=1,\ldots,p$,
	\begin{equation}
	\P\left\{ |y_j| \ge (t/\sqrt{p}) \|\x\|_2 \right\} \le 2\exp\left( -\eta t^2/2 \right)
	\end{equation}
	where $\eta=1$ for $\Hadamard$ a Hadamard matrix and $\eta=1/2$ for $\Hadamard$ a DCT matrix.
\end{thm}
As pointed out in \cite{ImprovedAnalysis}, the proof follows easily from Hoeffding's inequality (Thm.~\ref{thm:Hoeffding}). Since $\|\x\|_2=\|\y\|_2$, the theorem  implies that no single entry $y_j$ is likely to be far away from the average $\|\y\|_2/\sqrt{p}$.

Using the union bound, we derive the following results:
\begin{cor} \label{cor:FJLT}
	Let $\X$ be a $p\times n$ matrix with normalized columns, and $\Y=\Hadamard\Diag\X$ a random variable from the ROS applied to $\X$, then for all
 $\alpha\in(0,1)$, 
	\begin{align}
	\P\left\{ \|\Y\|_\text{max} \ge \frac{1}{\sqrt{p}}\cdot\sqrt{\frac{2}{\eta}\log\Big(\frac{2np}{\alpha}\Big)} \right\} &\le \alpha \label{eq:maxNorm} \\
	\P\left\{ \|\Y\|_\maxRow \ge  \sqrt{\frac{n}{p}}\cdot\sqrt{\frac{2}{\eta}\log\Big(\frac{2np}{\alpha}\Big)} \right\} &\le \alpha \label{eq:maxRowNorm} 
	\end{align}
\end{cor}
\begin{IEEEproof}
	Taking the union bound over all $np$ entries in the matrix $\Y$ gives (for each column of $\X$, we have $\|\x_i\|_2=1$)
	\[
	\P\left\{ \|\Y\|_\text{max} \ge t/\sqrt{p} \right\} \le 2np\exp\left( -\eta t^2/2 \right)
	\]
    which leads to \eqref{eq:maxNorm} if we choose $t^2=\frac{2}{\eta}\log(2np/\alpha)$.
	\newcommand{\Yi}{\Y_{j,:}}
	To derive the second equation, let $\Yi$ denote the $j$-th row of $\Y$. We apply the union bound to get
	\[
	\P\left\{ \|\Yi\|_\infty \ge t/\sqrt{p} \right\} \le 2n\exp\left( -\eta t^2/2 \right)
	\]
	and we bound the $\ell_2$ norm of $\Yi\in\R^{1\times n}$ with $\sqrt{n}$ times the $\ell_\infty$ norm,
	\begin{equation}\label{eq:RowNormY}
	\P\left\{ \|\Yi\|_2 \ge \sqrt{\frac{n}{p}}\cdot t \right\} \le 2n\exp\left( -\eta t^2/2 \right)
	\end{equation}
	from which \eqref{eq:maxRowNorm} follows by taking the union bound over $p$ rows and again choosing $t^2=\frac{2}{\eta}\log(2np/\alpha)$.
\end{IEEEproof}

	Note that for $\Hadamard$ a Hadamard matrix, the result in~\eqref{eq:RowNormY} shows the square of the $\ell_2$ norm of $\Y_{j,:}$ is unlikely to be larger than its mean value. To see this, 
    note that the $(u,l)$ entry of the Hadamard matrix  
     $H_{u,l}$ is either $\frac{1}{\sqrt{p}}$ or $-\frac{1}{\sqrt{p}}$, and $\Diag=\diag([D_1,\ldots,D_p])$ where each $D_i$ is $\pm1$ with equal probability. Without loss of generality, we consider the first row of $\Y$, $\Y_{1,:}$, and find the expectation of its $k$-th element squared:
	\begin{align}
	\E[Y_{1,k}^2] & =\E\Big[\Big(\sum_{l=1}^{p}D_lH_{1,l}X_{l,k}\Big)^2\Big]\nonumber\\
	&=\sum_{l=1}^{p}\E[D_l^2]H_{1,l}^2X_{l,k}^2\nonumber\\
	&+\sum_{l_1\neq l_2} \E[D_{l_1}D_{l_2}]H_{1,l_1}H_{1,l_2}X_{l_1,k}X_{l_2,k}\nonumber\\
	&=\frac{1}{p}\nonumber
	\end{align}
	since $\E[D_l]=0$, $\E[D_l^2]=1$, and $\X$ has normalized columns, i.e., $\sum_{l=1}^{p}X_{l,k}^2=1$. Thus, we get $\E[\|\Y_{1,:}\|_2^2]=\sum_{k=1}^{n}\E[Y_{1,k}^2]=\frac{n}{p}$. 

The importance of the preconditioning by the ROS prior to sub-sampling is the reduction of the norms over the worst-case values. As we will see in the subsequent sections, the variance of our sample mean and covariance estimators depends on the quantities such as the maximum absolute value of the entries and the maximum $\ell_2$ norm of the rows of the data matrix. Therefore, the preconditioning step \eqref{eq:ROS} is essential to obtain accurate and reliable estimates. For example, let us consider a data matrix $\X$ with normalized columns, then it is possible for $\|\X\|_\text{max}=1$, which would lead to 
weak bounds when inserted into our estimates. The best possible norm is $1/\sqrt{p}$ which occurs if all entries have the same magnitude. Our result in Corollary~\ref{cor:FJLT} states that, under the ROS, with high probability all the entries of $\Y=\Hadamard\Diag\X$ have comparable magnitude and it is unlikely $\|\Y\|_\text{max}$ is larger than $\sqrt{\log(np)}/\sqrt{p}$, which leads to lower variance and improved accuracy of our estimates.

Next, we present a result about the effect of sub-sampling on reduction of the Euclidean norm. Let $\w=\RR\RR^T\x$ denote the sub-sampled version of $\x\in\R^p$. Obviously, $\|\w\|_2^2 \leq \|\x\|_2^2$ and for data sets with a few large entries, $\|\w\|_2^2$ may be very close to $\|\x\|_2^2$ meaning that sub-sampling has not appreciably decreased the Euclidean norm. For example, consider a vector $\x=[1,0.1,0.01,0.001]^T$ where we wish to sample two out of four entries uniformly at random. In this case, $\|\w\|_2^2$ takes extreme values such that it might be very close to either $\|\x\|_2^2$ or zero. However, if we precondition the vector $\x$ prior to sub-sampling based on \eqref{eq:ROS}, this almost never happens, and the norm is reduced by nearly $m/p$ as one would hope for:
\begin{cor}[Norm bounds for ROS \eqref{eq:ROS}]\label{cor-rho}
		Let $\x\in\R^p$, and $\y=\Hadamard\Diag\x$ a random variable from the ROS applied to $\x$.  Define $\w\in\R^p$ to be a sampled version of $\y$, keeping $m$ of $p$ entries uniformly at random (without replacement). Then with probability greater than $1-\alpha$,
		\begin{equation}
		\|\w\|_2^2 \le \frac{m}{p} \frac{2}{\eta}\log\Big(\frac{2p}{\alpha}\Big) \|\x\|_2^2
		\end{equation}
		where $\eta=1$ for $\Hadamard$ a Hadamard matrix and $\eta=1/2$ for $\Hadamard$ a DCT matrix. Moreover, if $\{\w_i\}_{i=1}^{n}\in\R^p$ are sampled versions of the preconditioned data $\Y=\Hadamard\Diag\X$, with probability greater than $1-\alpha$,
		\begin{equation}
		\|\w_i\|_2^2 \le \frac{m}{p} \frac{2}{\eta}\log\Big(\frac{2np}{\alpha}\Big) \|\x_i\|_2^2,\; i=1,\ldots,n.
		\end{equation}
\end{cor}
\begin{IEEEproof}
		Regardless of how we sample, it holds deterministically that $\|\w\|_2 \le \sqrt{m}\|\y\|_\infty$ and this bound is reasonably sharp since the entries of $\y$ are  designed to have approximately the same magnitude. Using Thm.~\ref{thm:FJLT} and the union bound, the probability that $\|\y\|_\infty \ge (t/\sqrt{p}) \|\x\|_2$ is less than $2p e^{-\eta t^2/2}$. Then, we choose $t^2=\frac{2}{\eta}\log(2p/\alpha)$. Finally, we use the union bound when we have $n$ data samples. 
\end{IEEEproof}
As a result of Corollary~\ref{cor-rho}, when original data samples $\x_1,\ldots,\x_n$  are first preconditioned and then sub-sampled, by choosing $\alpha=1/100$ we see that $\|\w_i\|_2^2 \leq \frac{m}{p}\frac{2}{\eta}\log(200np) \|\x_i\|_2^2$, $i=1,\ldots,n$, with probability greater than $0.99$.

\section{The Sample Mean Estimator}\label{sec:center}
We show that a rescaled version of the sample mean of $\left\{ \mathbf{R}_{i}\mathbf{R}_{i}^{T}\mathbf{x}_{i}\right\} _{i=1}^{n}$, where $m$ out of $p$ entries of $\x_i$ are kept uniformly at random without replacement, is an unbiased estimator for the sample mean of the full data $\left\{ \mathbf{x}_{i}\right\} _{i=1}^{n}$. We will upper bound the error in both $\ell_{\infty}$ and $\ell_{2}$ norms, and show that these bounds are worse when the data set has a few large entries, which motivates our preconditioning.
\begin{thm} \label{thm:sampleMean}
	Let $\overline{\mathbf{x}}_n$ represent the sample mean of $\left\{ \mathbf{x}_{i}\right\} _{i=1}^{n}$, i.e., $\overline{\mathbf{x}}_n=\frac{1}{n}\sum_{i=1}^{n}\mathbf{x}_{i}$.
	Construct a rescaled version of the sample mean from $\left\{ \mathbf{R}_{i}\mathbf{R}_{i}^{T}\mathbf{x}_{i}\right\} _{i=1}^{n}$,
	where each column of $\mathbf{R}_{i}\in\mathbb{R}^{p\times m}$  is chosen uniformly at random from the set of all canonical basis vectors without replacement:
	\begin{equation}
		\widehat{\overline{\mathbf{x}}}_{n}=\frac{p}{m}\frac{1}{n}\sum_{i=1}^{n}\mathbf{R}_{i}\mathbf{R}_{i}^{T}\mathbf{x}_{i}.
	\end{equation}
	Then, $\widehat{\overline{\mathbf{x}}}_{n}$ is an unbiased estimator
	for $\overline{\mathbf{x}}_n$,
	i.e., $\mathbb{E}[\widehat{\overline{\mathbf{x}}}_{n}]=\overline{\mathbf{x}}_n$.
    Moreover, 	defining $\tau(m,p)$ as follows:
    \begin{equation}\label{eq:tau-m-p}
    \tau(m,p) \!:=\!\! \max\!\left\{\!\Big(\frac{p}{m}\!-\!1\Big)\!,\!1\right\} \!=\!
    \begin{cases} 
    (\frac{p}{m}\!-\!1) \!\!\!& \text{if } \frac{m}{p}\!\leq\! 0.5\\
    1       \!\!\!& \text{if } \frac{m}{p}\! >\! 0.5
    \end{cases}
    \end{equation}
	then for all $t\geq0$, with probability at least $1-\delta_{1}$, 
	\begin{equation}
	\delta_{1}=2p\exp\left(\frac{-nt^{2}/2}{(\frac{p}{m}-1)\nicefrac{\|\X\|_\maxRow^{2}}{n}+\nicefrac{\tau(m,p)\left\Vert \mathbf{X}\right\Vert_\text{max}t}{3}    }\right)\label{eq:fail_prob_sample_mean}
	\end{equation}
	 we have the following $\ell_{\infty}$ result:
	 \begin{equation}
	 \Vert \widehat{\overline{\mathbf{x}}}_{n}-\overline{\mathbf{x}}_n\Vert _{\infty}\leq t.\label{eq:error_sample_mean}
	 \end{equation}
\end{thm}
\begin{IEEEproof}
	First, we show that $\widehat{\overline{\mathbf{x}}}_{n}$ is an unbiased
	estimator:
	\[
	\mathbb{E}[\widehat{\overline{\mathbf{x}}}_{n}]=\frac{p}{m}\frac{1}{n}\sum_{i=1}^{n}\mathbb{E}\left[\mathbf{R}_{i}\mathbf{R}_{i}^{T}\right]\mathbf{x}_{i}=\frac{1}{n}\sum_{i=1}^{n}\mathbf{x}_{i}=\overline{\mathbf{x}}_n
	\]
	where we used Thm.~\ref{thm:Properties_Sketching}. To upper bound the error, we define:
	\begin{equation}
	\mathbf{u}:=\widehat{\overline{\mathbf{x}}}_{n}-\overline{\mathbf{x}}_n=\sum_{i=1}^{n}\frac{1}{n}\Big(\frac{p}{m}\mathbf{R}_{i}\mathbf{R}_{i}^{T}\mathbf{x}_{i}-\mathbf{x}_{i}\Big)
	\end{equation}
	and, thus, the $j$-th entry of $\mathbf{u}=[u_{1},\ldots,u_{p}]^{T}$
	can be written as a sum of independent centered random variables:
	\begin{equation}
	u_{j}=\sum_{i=1}^{n}z_{i},\;\text{where}\; z_{i}=\frac{1}{n}\mathbf{e}_{j}^{T}\Big(\frac{p}{m}\mathbf{R}_{i}\mathbf{R}_{i}^{T}\mathbf{x}_{i}-\mathbf{x}_{i}\Big).
	\end{equation}
	We now use the Bernstein inequality (Thm.~\ref{thm:Bernstein}) to show each entry of $\mathbf{u}$ is concentrated around zero with high probability. To do this, let us define $\tau(m,p):=\max\{(\frac{p}{m}-1),1\}$. We observe that each $z_{i}$ is bounded:
	\begin{equation}
	\left|z_{i}\right|\leq\frac{1}{n}\left\Vert \frac{p}{m}\mathbf{R}_{i}\mathbf{R}_{i}^{T}\mathbf{x}_{i}-\mathbf{x}_{i}\right\Vert _{\infty}\!\!\!\leq\frac{1}{n}\tau(m,p)\left\Vert \mathbf{X}\right\Vert_\text{max}
	\end{equation}
	since $\mathbf{R}_{i}\mathbf{R}_{i}^{T}\mathbf{x}_{i}$
	is a vector with $m$ entries of $\mathbf{x}_{i}$ and the rest equal to zero. Next, we find the variance of $z_i$, $\mathbb{E}[z_{i}^{2}]=(\nicefrac{1}{n^{2}})\mathbf{e}_{j}^{T}\boldsymbol{\Lambda}\mathbf{e}_{j}$, where we use Thm.~\ref{thm:Properties_Sketching} to compute $\boldsymbol{\Lambda}$:
	\begin{eqnarray*}
		\hspace{-7mm}& \boldsymbol{\Lambda}\hspace{-2mm} & =\mathbb{E}\Big[\Big(\frac{p}{m}\mathbf{R}_{i}\mathbf{R}_{i}^{T}\mathbf{x}_{i}-\mathbf{x}_{i}\Big)\Big(\frac{p}{m}\mathbf{R}_{i}\mathbf{R}_{i}^{T}\mathbf{x}_{i}-\mathbf{x}_{i}\Big)^{T}\Big]\\
		\hspace{-7mm}&  & =\frac{p^{2}}{m^{2}}\mathbb{E}\Big[\mathbf{R}_{i}\mathbf{R}_{i}^{T}\mathbf{x}_{i}\mathbf{x}_{i}^{T}\mathbf{R}_{i}\mathbf{R}_{i}^{T}\Big]-\mathbf{x}_{i}\mathbf{x}_{i}^{T}\\
		\hspace{-7mm}&  & =\frac{-\left(p-m\right)}{m\left(p-1\right)}\mathbf{x}_{i}\mathbf{x}_{i}^{T}+\frac{p\left(p-m\right)}{m\left(p-1\right)}\text{diag}\left(\mathbf{x}_{i}\mathbf{x}_{i}^{T}\right).
	\end{eqnarray*}
	Thus, $\mathbb{E}[z_{i}^{2}]=(\nicefrac{1}{n^{2}})(\frac{p}{m}-1)(\mathbf{e}_{j}^{T}\mathbf{x}_{i})^{2}$, and $\sigma^{2}=\sum_{i=1}^{n}\mathbb{E}[z_{i}^{2}]\leq(\nicefrac{1}{n^2})(\frac{p}{m}-1)\|\X\|_\maxRow^{2}$.
	Using the Bernstein inequality:
	\[
	\P\left\{ \left|u_{j}\right|\geq t\right\} \leq2\exp\left(\frac{-nt^{2}/2}{(\frac{p}{m}-1)\nicefrac{\|\X\|_\maxRow^{2}}{n}+\nicefrac{\tau(m,p)\left\Vert \mathbf{X}\right\Vert_\text{max}t}{3}    }\right).
	\]
	Finally, we use the union bound over all $p$ entries of $\mathbf{u}$.
\end{IEEEproof}
\begin{remark}
The result in Thm.~\ref{thm:sampleMean} can be used to upper bound the error of our sample mean estimator $\widehat{\overline{\mathbf{x}}}_{n}$ in $\ell_2$ norm. For all $t\geq0$, with probability at least $1-\delta_1$, where $\delta_1$ defined in~\eqref{eq:fail_prob_sample_mean}, we have: 
\begin{equation}
\frac{1}{\sqrt{p}}\Vert\widehat{\overline{\mathbf{x}}}_{n}-\overline{\mathbf{x}}_n\Vert _{2}\leq t.
\end{equation}
\end{remark}
\begin{remark}
Solving for $t$ in \eqref{eq:fail_prob_sample_mean} gives the following expression in terms of failure probability $\delta_1$:
\newcommand{\logg}{\log\left( \frac{2p}{\delta_1}\right)}
\begin{equation}
t = \frac{1}{n}\left( \frac{\tau(m,p)}{3}\|\X\|_\text{max} \logg 
+ \sqrt{ \left( \frac{\tau(m,p)}{3}\|\X\|_\text{max} \logg \right)^2
    + 2\left(\frac{p}{m}-1\right)\logg \|\X\|_\maxRow^{2}
}
\right)
\end{equation}    
\end{remark}

\begin{figure}
	\centering
	\includegraphics[width=\figWidth]{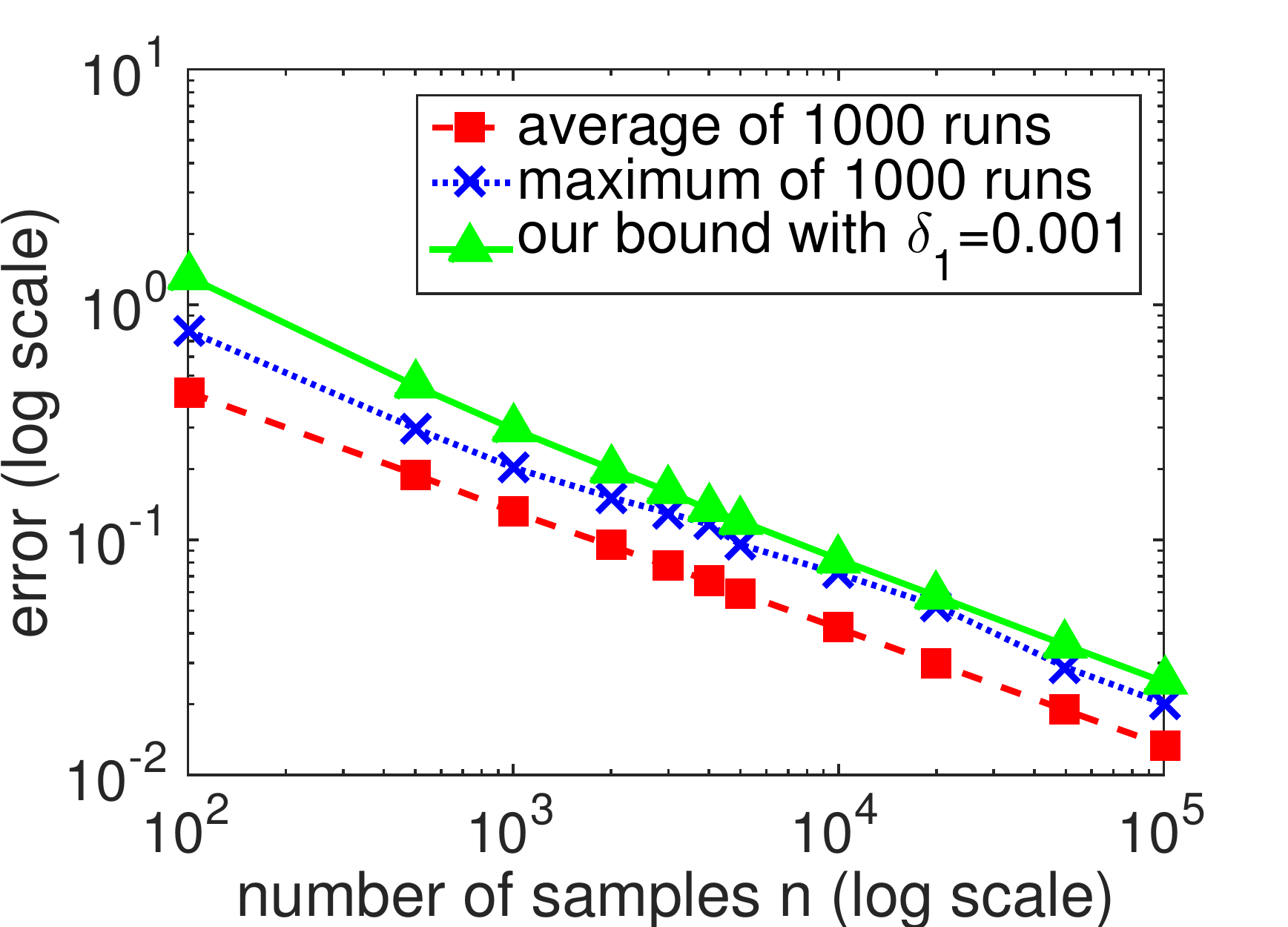}
	\caption{Verifying the sharpness of Thm.~\ref{thm:sampleMean} on the synthetic data. For each $n$ 
we report the average and maximum of the sample mean estimation error in $1000$ runs compared with the theoretical error bound when $\delta_1=0.001$. The theoretical error bound is tight and decays exponentially as $n$ increases. 
		\label{fig:convergence_sample_mean}
	}
\end{figure}

We consider a numerical experiment on synthetic data set to show the precision of Thm.~\ref{thm:sampleMean}. We set the parameters $p=100$ and compression factor $\gamma=m/p=0.3$ and consider $1000$ runs for different values of $n$.  Let $ \mathcal{N}(\boldsymbol{\mu},\boldsymbol{\Sigma})$ denote a multivariate Gaussian distribution parameterized by its mean $\boldsymbol{\mu}$ and covariance matrix $\boldsymbol{\Sigma}$. In each run, we generate a set of $n$ samples in $\R^{p}$ from the probabilistic generative model $\x_i=\overline{\x}+\epsilon_i$, where $\overline{\x}$ is a fixed vector drawn from the Gaussian distribution and the additive noise $\epsilon_i$ is drawn i.i.d.~from $\mathcal{N}(\boldsymbol{0},\mathbf{I}_p)$. We then keep $m=30$ entries from each data sample uniformly at random without replacement to obtain a sparse matrix. Using Thm.~\ref{thm:sampleMean}, we find the estimates of the sample mean from the sparse matrix and 
compare with the actual
$\ell_\infty$ error between the estimates and true sample means. Fig.~\ref{fig:convergence_sample_mean} reports the average and maximum of $1000$ runs for each value of $n$ and compares that with the theoretical error bound $t$ in~\eqref{eq:fail_prob_sample_mean} obtained by setting the failure probability $\delta_1=0.001$. The theoretical error bound is quite tight since it is close to the maximum of $1000$ runs.

In Thm.~\ref{thm:sampleMean}, the failure probability $\delta_1$ depends on the properties of the data set such as the maximum absolute value of $\X$.
For a data matrix $\X$ with normalized columns, values of $\|\X\|_\text{max}$ and $\|\X\|_\maxRow^{2}$ can be relatively large due to the existence of large entries in $\X$ and, in the extreme case, we can get $\|\X\|_\text{max}=1$ and $\|\X\|_\maxRow^{2}=n$. Since both $\|\X\|_\text{max}$ and $\|\X\|_\maxRow^{2}$ are in the denominator, large values of these quantities work against the accuracy of the estimator and makes the failure probability $\delta_1$ closer to $1$. This fact motivates the use of preconditioning transformation discussed in Section~\ref{sec:precondition} to smooth out large entries of $\X$ and reduce the values of $\|\X\|_\text{max}$ and $\|\X\|_\maxRow^{2}$. 
\begin{cor}\label{cor:sampleMean}
In the setting of Thm.~\ref{thm:sampleMean}, assume $\X$ is preconditioned by the ROS~\eqref{eq:ROS} and $\gamma=m/p\leq0.5$.
By using the results of Corollary~\ref{cor:FJLT}, we can find an upper bound for the failure probability $\delta_1$ that holds with probability greater than $0.99$:
\begin{equation}
\delta_1\leq 2p\exp\left(\frac{-mnt^2}{\nicefrac{4}{\eta}\log(200np)(1+\nicefrac{\sqrt{p}t}{3})}\right),
\end{equation}
and, thus, we can achieve high accuracy, e.g.,~$\delta_1\leq0.001$, for
\begin{equation}
m\geq\frac{1}{n}\cdot \frac{4}{\eta}\log(200np)\log(2000p)(t^{-2}+\frac{\sqrt{p}t^{-1}}{3})\label{eq:lower-bound-m}
\end{equation}
where $\eta=1$ for $\Hadamard$ a Hadamard matrix and $\eta=1/2$ for $\Hadamard$ a DCT matrix. Recall that $t$  is the upper bound for estimation error in~\eqref{eq:error_sample_mean}. To gain intuition and provide indicative values of $m$, we set $p=512$, $\eta=1$, and $t=0.01$.  Then, for example, the values of the lower bound in~\eqref{eq:lower-bound-m} are $137.2$, $15.1$, and $1.6$ for $n=10^5$, $10^6$, and $10^7$ respectively. 
Since $m$ should be a natural number, we need to sample $m=138$, $m=16$, and $m=2$ entries per data sample. This means that as the number of samples $n$ increases, we can sample fewer entries per data sample, which makes our approach applicable to large-scale data sets.

To be formal, as $n$ grows with $p$ and $t$ fixed, if the number of sub-sampled entries per data sample $m$ is proportional to $\order(\log n/n)$, our sample mean estimator is accurate with high probability. Therefore, the amount of data we need to keep increases like $mn\propto \order(\log n)$.
\end{cor}
\section{The Covariance Estimator}\label{sec:covariance}
\newcommand{\Cemp}{\mathbf{C}_\text{emp}}
\newcommand{\Cemph}{\widehat{\mathbf{C}}_\text{emp}}
\newcommand{\Cn}{\widehat{\mathbf{C}}_n}
In this section, we study the problem of covariance estimation for a set of data samples $\left\{\mathbf{x}_{i}\right\} _{i=1}^{n}$ from $\left\{ \mathbf{R}_{i}\mathbf{R}_{i}^{T}\mathbf{x}_{i}\right\} _{i=1}^{n}$, where $m$ out of $p$ entries of each $\x_i$ are kept uniformly at random without replacement. We propose an unbiased estimator for the covariance matrix of the full data $\Cemp=\nicefrac{1}{n}\sum_{i=1}^{n}\x_i\x_i^T$ and study the closeness of our proposed covariance estimator to $\Cemp$ based on the spectral norm. 
Recall
we do not impose structural assumptions on the covariance matrix such as $\Cemp$ being low-rank. 

To begin, consider a rescaled version of the empirical covariance matrix of the sub-sampled data $\left\{ \mathbf{R}_{i}\mathbf{R}_{i}^{T}\mathbf{x}_{i}\right\} _{i=1}^{n}$:
\begin{equation}
\Cemph:=\frac{p(p-1)}{m(m-1)}\cdot\frac{1}{n}\sum_{i=1}^{n}\RR_i\RR_i^{T}\x_i\x_i^{T}\RR_i\RR_i^{T}.
\end{equation}
Based on Thm.~\ref{thm:Properties_Sketching}, we can compute the expectation of $\Cemph$:
\begin{equation}
\E[\Cemph]=\Cemp+\frac{(p-m)}{(m-1)}\diag(\Cemp)
\end{equation}
which consists of two terms, the covariance matrix of the full data $\Cemp$ (desired term) and an additional bias term that contains the elements on the main diagonal of $\Cemp$. However, as in~\cite{CovEstimationCSAarti}, we can easily modify $\Cemph$ to obtain an unbiased estimator:
\begin{equation}
\Cn:=\Cemph-\frac{(p-m)}{(p-1)}\diag(\Cemph)
\end{equation}
where revisiting Thm.~\ref{thm:Properties_Sketching} shows that $\Cn$ is an unbiased estimator for $\Cemp$, i.e.,~$\E[\Cn]=\Cemp$. Next, we present a theorem to show the closeness of our proposed estimator $\Cn$ to the covariance matrix $\Cemp$.

Before stating the result, let us define $\w_i:=\RR_i\RR_i^{T}\x_i$, $i=1,\ldots,n$, which is an $m$-sparse vector containing $m$ entries of $\x_i$. We introduce a parameter $\rho>0$ such that for all $i=1,\ldots,n$ we have $\|\w_i\|_2^2 \leq \rho \|\x_i\|_2^2$. Obviously, $\|\w_i\|_2^2 \leq \|\x_i\|_2^2$ and for data sets with a few large entries, we can have $\|\w_i\|_2^2 = \|\x_i\|_2^2$ meaning that sub-sampling has not decreased the Euclidean norm. Therefore, $\rho\leq1$ and we can always take $\rho=1$. However, if we first apply the preconditioning transformation $\Hadamard\Diag$~\eqref{eq:ROS} to the data, we see that, with high probability, the sub-sampling operation decreases the Euclidean norm by a factor proportional to the compression factor $\gamma=m/p$. In fact, Corollary~\ref{cor-rho} with $\alpha=1/100$ shows that $\rho=\frac{m}{p}\frac{2}{\eta}\log(200np)$ with probability greater than $0.99$. As we will see, this is of great importance to decrease the variance of our covariance estimator and achieve high accuracy, which motivates using the preconditioning transformation before sub-sampling.

\begin{figure}
	\begin{centering}
		\subfloat[]{\begin{centering}
        		\includegraphics[width=\figWidth]{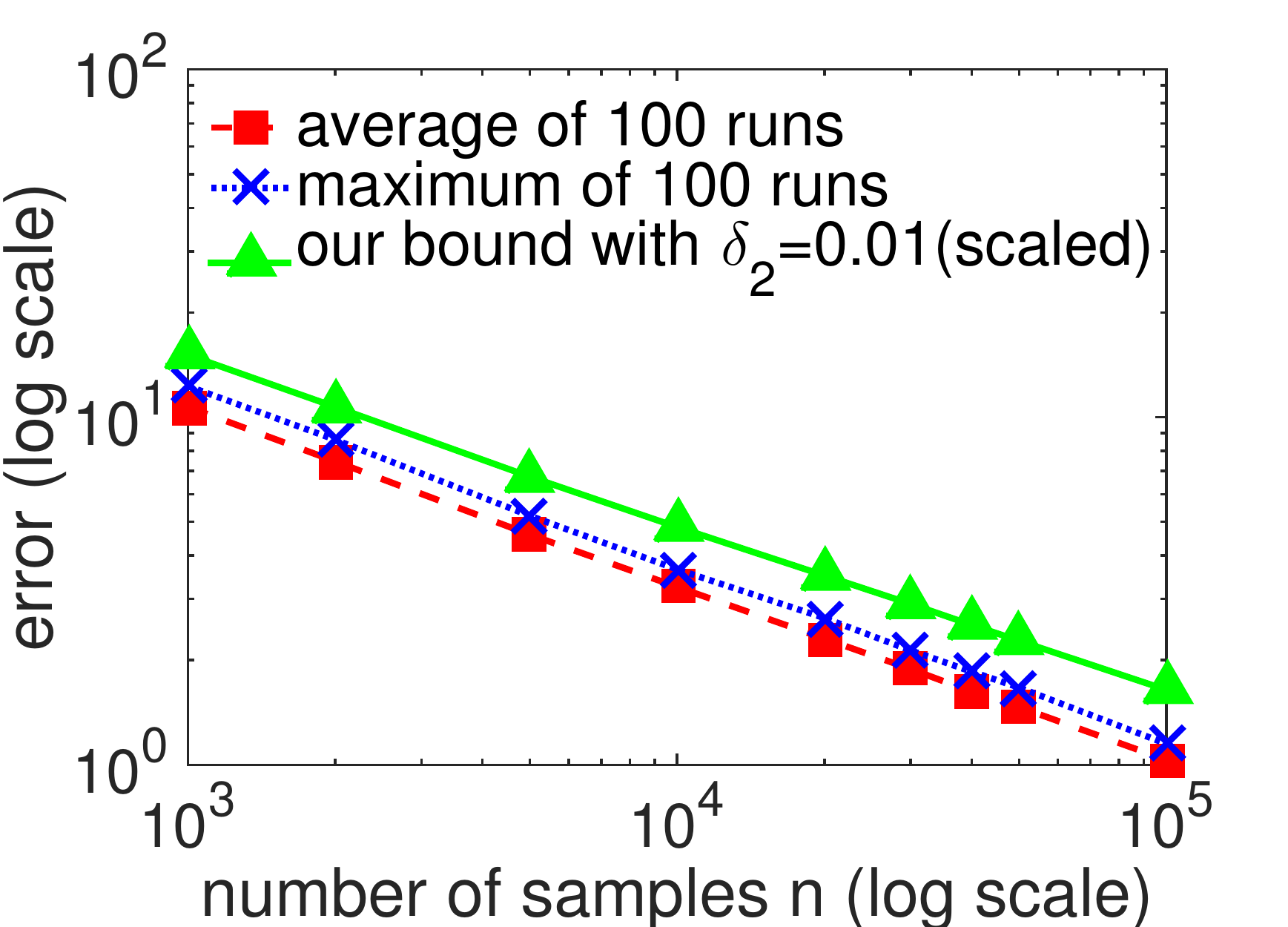}
				\par\end{centering}
			
		}\subfloat[]{\begin{centering}
			\includegraphics[width=\figWidth]{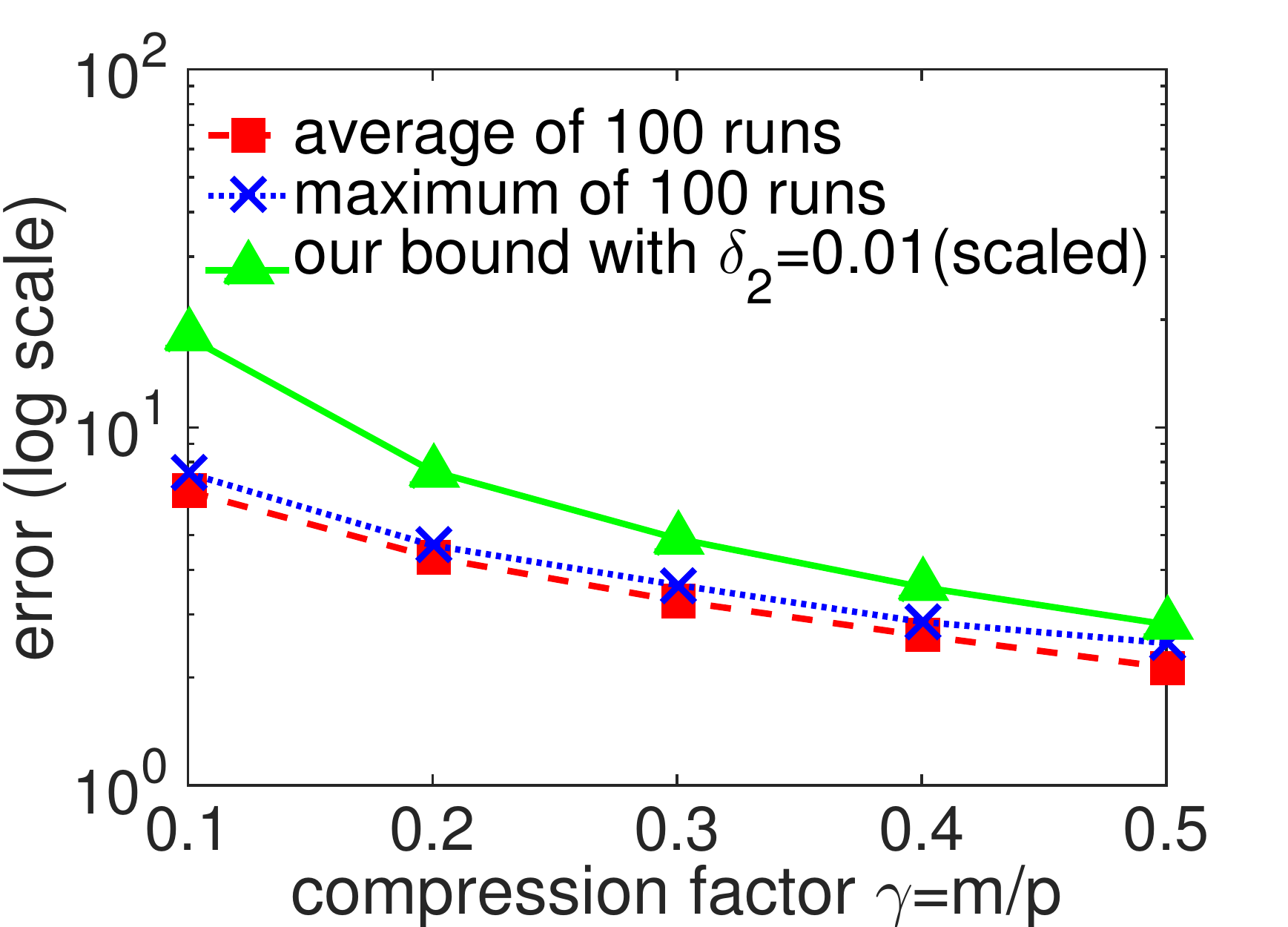}
			\par\end{centering}
	}
	\par\end{centering}
\caption{Verifying the accuracy of Thm.~\ref{thm:covariance} on synthetic data. We set $p=1000$ and plot the average and maximum of covariance estimation error in $100$ runs for (a) varying $n$ when $\gamma=0.3$ fixed, and (b) varying $\gamma$ when $n=10p$ fixed. The empirical values are compared with the theoretical error bound for $\delta_2=0.01$ and scaled by a factor of $10$. 
Our bounds are accurate to within an order of magnitude and they are representative of the empirical behavior of our covariance estimator in terms of $n$ and $\gamma$. 
	\label{fig:convergence_covariance}}
\end{figure}

\begin{thm}\label{thm:covariance}
Let $\Cemp$ represent the covariance matrix of $\left\{ \mathbf{x}_{i}\right\} _{i=1}^{n}$ and construct a rescaled version of the empirical covariance matrix from $\left\{\w_i= \mathbf{R}_{i}\mathbf{R}_{i}^{T}\mathbf{x}_{i}\right\} _{i=1}^{n}$,
where each column of $\mathbf{R}_{i}\in\mathbb{R}^{p\times m}$  is chosen uniformly at random from the set of all canonical basis vectors without replacement:
\begin{equation}
\Cemph=\frac{p(p-1)}{m(m-1)}\cdot\frac{1}{n}\sum_{i=1}^{n}\RR_i\RR_i^{T}\x_i\x_i^{T}\RR_i\RR_i^{T}.
\end{equation}
Let $\rho>0$ be a bound 
such that for all $i=1,\ldots,n$, we have $\|\w_i\|_2^2 \leq \rho \|\x_i\|_2^2$
(in particular, we can always take $\rho=1$). Then,
\begin{equation}
\Cn=\Cemph-\frac{(p-m)}{(p-1)}\diag(\Cemph)
\end{equation}
is an unbiased estimator for $\Cemp$, and for all $t\geq 0$,
\begin{equation}
\P \left\{\|\Cn-\Cemp\|_2\leq t \right\}\geq 1-\delta_2,\;\;\delta_2=p\exp\left(\frac{-\nicefrac{t^2}{2}}{\sigma^2+\nicefrac{Lt}{3}}\right)\label{eq:cov-converge-delta2}
\end{equation}
where 
\begin{equation}
L=\hspace{-1mm} \frac{1}{n} \left\{ \left(\frac{p(p-1)}{m(m-1)}\rho+1\right)\|\X\|_\maxCol^2 + \frac{p(p-m)}{m(m-1)}\|\X\|_\text{max}^2\right\}\label{eq:covariance-L}
\end{equation}
and $\sigma^2=\|\E[(\Cn-\Cemp)^2]\|_2$ represents the variance:
\begin{eqnarray}
&\hspace{-9mm}\sigma^2&\hspace{-5mm}\leq \frac{1}{n}  \left\{\left(\frac{p(p-1)}{m(m-1)}\rho-1\right)\|\X\|_\maxCol^2\|\Cemp\|_2    \right.   \nonumber \\
&&\hspace{-5mm} \left. +\frac{p(p-1)(p-m)}{m(m-1)^2}\rho \|\X\|_\maxCol^2\|\diag(\Cemp)\|_2  \right.  \nonumber \\
&&\hspace{-5mm} +\frac{2p(p-1)(p-m)}{m(m-1)^2}\|\X\|_\text{max}^2\frac{\|\X\|_F^2}{n} \nonumber \\
&&\hspace{-5mm} \left. +\frac{p(p-m)^2}{m(m-1)^2}\frac{\max_{j=1,\ldots,p}\sum_{i=1}^{n}X_{j,i}^4}{n}     \right\}.\label{eq:covariance_sigma}
\end{eqnarray}
\end{thm}
\begin{IEEEproof}
The proof follows from the matrix Bernstein inequality (Thm.~\ref{thm:MatrixBernstein}) and delayed to Appendix~\ref{sec:proof-of-covarinace}. 
\end{IEEEproof}
To gain some intuition and verify the accuracy of Thm.~\ref{thm:covariance}, we consider a numerical experiment. We set the parameter $p=1000$ and show the accuracy of our proposed covariance estimator $\Cn$ for various numbers of samples $n$ and compression factors $\gamma$. Fig.~\ref{fig:convergence_covariance}(a) shows the closeness of our covariance estimator $\Cn$ to the covariance matrix $\Cemp$, i.e., $\|\Cn-\Cemp\|_2$, over $100$ runs for various values of $n$ when $\gamma=m/p=0.3$ is fixed. In each run, we generate a set of $n$ data samples using the probabilistic generative model $\x_i=\sum_{j=1}^{k}\kappa_{ij}\lambda_{j}\mathbf{u}_j$, where we set $k=5$ and $\mathbf{U}=[\mathbf{u}_1,\ldots,\mathbf{u}_5]$ is a matrix of principal components with orthonormal columns obtained by performing QR decomposition on a $p\times k$ matrix with i.i.d.~entries from $\mathcal{N}(0,1)$. The coefficients $\kappa_{ij}$ are drawn i.i.d.~from $\mathcal{N}(0,1)$ and the vector $\boldsymbol{\lambda}$ represents the energy of the data in each principal direction and we choose $\boldsymbol{\lambda}=(10,8,6,4,2)$. The empirical value of estimation error $\|\Cn-\Cemp\|_2$ is compared with the theoretical error bound $t$ in \eqref{eq:cov-converge-delta2} when the failure probability is chosen $\delta_2=0.01$, and the resulting error bound is scaled by a factor of $10$.

Furthermore, we plot the empirical value of the estimation error vs.~compression factor $\gamma$ in Fig.~\ref{fig:convergence_covariance}(b) over $100$ runs when a fixed number of samples $n=10p$ are generated using the same generative model. As before, the empirical value is compared with the theoretical error bound when we choose $\delta_2=0.01$ and our bound is scaled by the same factor of $10$. 
Our bounds are accurate to within an order of magnitude, and the theoretical result in Thm.~\ref{thm:covariance} correctly captures the dependence of the  estimation error $\|\Cn-\Cemp\|_2$ in terms of the parameters $n$ and $\gamma$. 
For example, as $n$ increases, the estimation error decreases exponentially for a fixed compression factor $\gamma$. 

The other important consequence of Thm.~\ref{thm:covariance} is revealing the connections between accuracy of our covariance estimator $\Cn$ and some properties of the data set $\X$ as well as the compression factor $\gamma$. Note that large values of parameters $L$ and $\sigma^2$ work against the accuracy of $\Cn$ and make the failure probability $\delta_2$ closer to $1$, since they are in the denominator in \eqref{eq:cov-converge-delta2}. 
Let us assume that $\X$ has normalized columns so that $\|\X\|_\maxCol=1$ and $\|\X\|_F^2=n$. 
With this normalization, $\|\Cemp\|_2\le1$, and $\|\diag(\Cemp)\|_2 \le 1$ as well, which follows from exercise 27 in \S3.3 \cite{HornJohnson2} and $\Cemp \succeq 0$. 
In this case, both $L$ and $\sigma^2$ scale as $\frac{1}{n}$ and the estimation error decreases exponentially as $n$ increases for a fixed compression factor. Moreover, for a fixed $n$, $L\propto\order(\frac{p^2}{m^2})$ and $\sigma^2\propto\order(\frac{p^3}{m^3})$. 
However, if we precondition the data $\X$ before sub-sampling as discussed in Section~\ref{sec:precondition}, then $\rho\propto\order(\frac{m}{p})$ and the maximum absolute value of the entries $\|\X\|_\text{max}$ of the preconditioned data is proportional to $\frac{1}{\sqrt{p}}$ ignoring logarithmic factors. Thus, the leading term in $L$ scales as $\order(\frac{p}{m})$ and the leading term in $\sigma^2$ scales as $\order(\frac{p^2}{m^2})$ which means that they are both reduced by a factor of $\frac{m}{p}$ under the preconditioning transformation.  
Specifically, simplifying \eqref{eq:covariance_sigma} by dropping lower-order terms, assuming $p\gg m\gg 1$,  using the normalization of $\X$ discussed above, as well as assuming preconditioning so $\rho=m/p$ and $\|\X\|_\text{max}^2\approx 1/p$, gives
\begin{equation}\label{eq:simplified}
\sigma^2 \lesssim \frac{1}{n}  \left\{ 
   \frac{p}{m}\|\Cemp\|_2  
  +\frac{p^2}{m^2} \|\diag(\Cemp)\|_2  
  +\frac{2p^2}{m^3}
  +\frac{p}{m^3} 
\right\}.
\end{equation}

Using just $\|\Cemp\|$, $\|\diag(\Cemp)\| \le 1$ then gives the bound $\sigma^2 \lesssim\mathcal{O}\left( \frac{1}{n}\frac{p^2}{m^2} \right)$, but this can be improved in special cases. 
Based on \eqref{eq:simplified}, 
we now consider tighter bounds for the few special cases, still assuming the data have been preconditioned so that $\Cemp = \frac{1}{n} \Hadamard\Diag \X\X^T\Diag^T\Hadamard^T$:
\begin{enumerate}
\item If each $\xi$ is a canonical basis vector chosen uniformly-at-random, then $\Cemp = \diag(\Cemp)= p^{-1} \eye_p$ in the limit as $n\rightarrow\infty$. The third term in \eqref{eq:simplified} dominates, and the bound is $1/n\cdot p^2/m^3$ which is quite strong.
\item If each entry of $\X$ is chosen i.i.d.\ $\mathcal{N}( 0, 1/p )$, then the columns have unit norm in expectation and again $\Cemp \rightarrow p^{-1} \eye_p$ as $n\rightarrow \infty$, so the bound is the same as the previous case (and preconditioning neither helps nor hurts).
\item If $\xi = \x$ for all columns $i=1,\ldots,n$ and for some fixed (and normalized) column $\x$, then $\Cemp = \Hadamard\Diag\x\x^T\Diag^T\Hadamard^T$, so $\|\Cemp\|_2 = 1$. 
    For example, if $\x$ is a canonical basis vector, then $\Cemp$ is the outer-product of a column of the Hadamard matrix, so $\diag(\Cemp) = p^{-1} \eye_p$ and thus $\sigma^2\lesssim 1/n \cdot p/m$, similar to \cite{CovEstimationCSAarti}. As well as improving $\rho$ and $\|\X\|_\text{max}$, in this case preconditioning has the effect of spreading out the energy in $\Cemp$ away from the diagonal. Without preconditioning, then  $\Cemp = \x\x^T$, so if $\x$ is a standard basis vector, $\Cemp=\diag(\Cemp)$. Intuitively, this is a bad case for non-preconditioned sampling, since there is a slim chance of sampling the only non-zero coordinate. Note that this perfectly correlated case is difficult for approaches based on $\rand \X$ since each column is the same so the measurements are redundant, whereas our approach uses a unique sampling matrix for each column $i$ and thus gives more information.   
\end{enumerate}  
   
In order to clarify the discussion, we provide a simple numerical experiment on a synthetic data set with a few large entries. We then show the effectiveness of the preconditioning transformation $\Hadamard\Diag$ on the accuracy of our covariance estimator $\Cn$ and, consequently, accuracy of the estimated principal components on this data set. 
 \begin{figure}
 	\centering
 	\includegraphics[width=\figWidth]{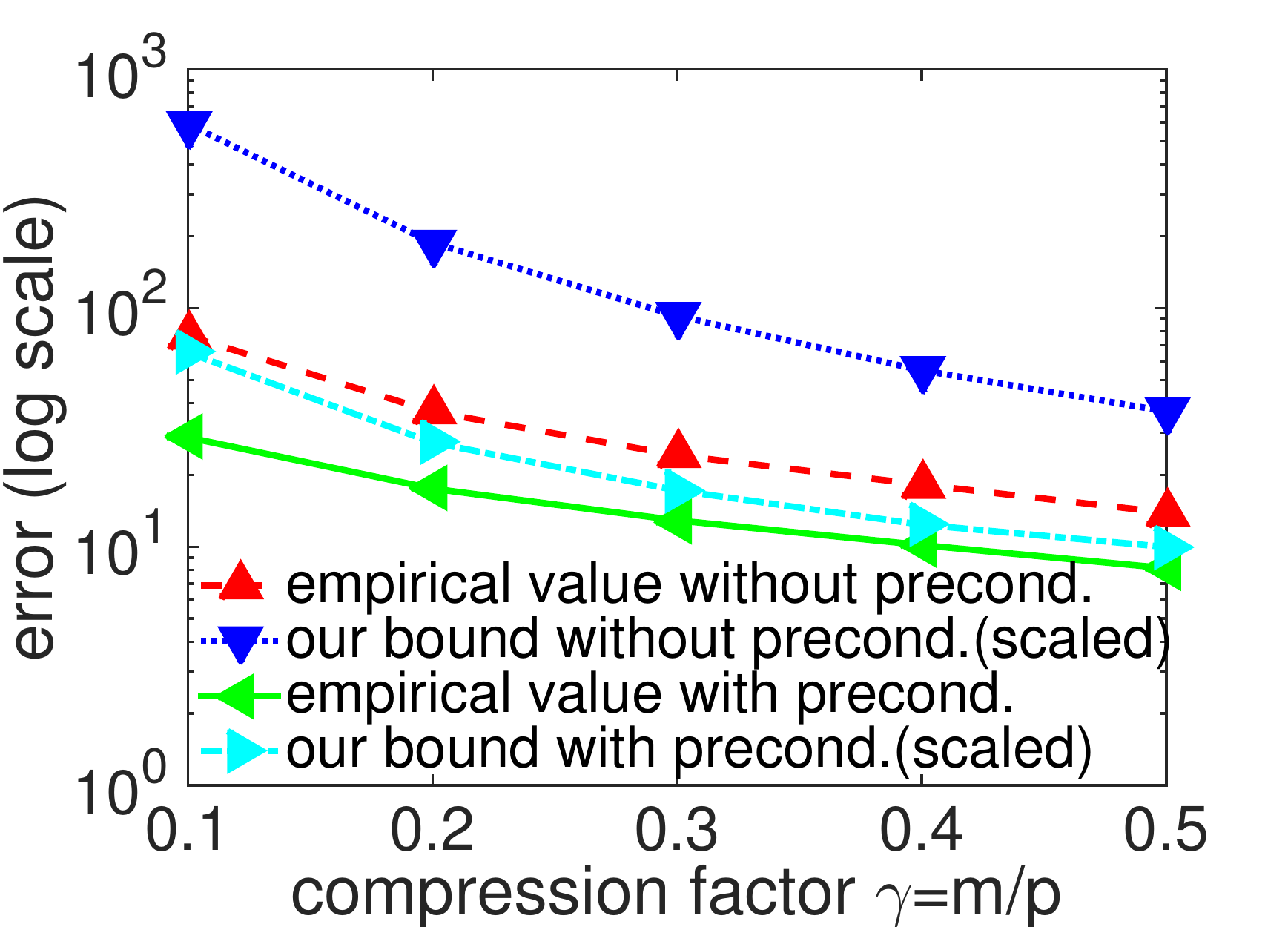}
 	\caption{Effectiveness of preconditioning on the synthetic data set with few large entries. We plot the average of covariance estimation error over $100$ runs for varying $\gamma$ when $p=512$ and $n=1024$ in two cases of sub-sampling $\X$ (without preconditioning) and $\Y=\Hadamard\Diag\X$ (with preconditioning). We compare the empirical values with the theoretical error bound for $\delta_2=0.01$ and scaled by a factor $10$ in these two cases. 
    The preconditioning transformation $\Hadamard\Diag$ leads to a noticeable reduction of estimation error in both empirical and theoretical results. 
 		\label{fig:convergence-cov-precond}
 	}
 \end{figure}
 
In our experiment, we set $p=512$, and generate $n=1024$ data samples from the same probabilistic generative model with $k=10$ and $\mathbf{U}\in\R^{p\times k}$ containing $10$ principal components chosen from the set of all canonical basis vectors, and $\boldsymbol{\lambda}=(10,9,\ldots,1)$. In Fig.~\ref{fig:convergence-cov-precond}, we plot the average of estimation error $\|\Cn-\Cemp\|_2$ over $100$ runs for various values of the compression factor $\gamma=m/p$.
 As described in Section \ref{sec:precondition}, we can use a simple unitary transformation $\Hadamard\Diag$ to precondition the data and smooth out large entries. Thus, we consider the case where the data is first preconditioned, i.e., $\Y=\Hadamard\Diag\X$. 
 In this case the estimation error is $\|\Cn-\Cemp\|_2$, where $\Cemp=\frac{1}{n}\Y\Y^{T}$, and is also plotted in Fig.~\ref{fig:convergence-cov-precond}. Moreover, we report the theoretical error bounds when we choose $
 \delta_2=0.01$ and scale our bounds by the same factor of $10$. The preconditioning decreases the error by almost a factor of $2$, both in experiment and via the theoretical bounds. 

To show the importance of this error reduction in covariance estimation, we find the number of recovered principal components obtained from the eigendecomposition of $\Cn$ for both cases. After finding the first $k=10$ eigenvectors of $\Cn$, we compute the inner product magnitude between the recovered and true principal components and we declare that a principal component is ``recovered'' if the corresponding inner product magnitude is greater than $0.95$. The mean and standard deviation of the number of recovered principal components for varying compression factors $\gamma=m/p$ are reported in Table~\ref{num-recovered-precon}. As we see, the preconditioning transformation leads to a significant gain in accuracy of the estimated principal components, especially for small values of the compression factor, which are of great importance for the big data and distributed data settings.
Additionally, the variance in the estimate is much reduced across the whole range of $\gamma$.
\begin{table}[!t]
	\caption{Number of recovered principal components for various values of compression factor $\gamma=m/p$ over $100$ runs}
	\label{num-recovered-precon}
    \centering
	\begin{tabular}{lcccc}
		\toprule
        & \multicolumn{2}{l}{(without preconditioning)} & \multicolumn{2}{l}{(with preconditioning)} \\
        \cmidrule(rl){2-3}
        \cmidrule(l){4-5}
		$\gamma$ & mean & standard deviation & mean & standard deviation \\
		\midrule
		0.1 & 0.98 & 0.99 & 5.12 & 0.40\\
		0.2 & 3.53 & 1.76 & 7.01 & 0.10 \\
		0.3 & 6.85 & 1.67 & 8.00 & 0\hphantom{.10}\\
		0.4 & 8.18 & 1.58 & 8.42 & 0.49\\
		0.5 & 9.31 & 1.03 & 9.00 & 0\hphantom{.10}\\
		\bottomrule
	\end{tabular} 
\end{table}

\section{Sparsified K-means Clustering}\label{sec:CK-means}
Clustering is a commonly used unsupervised learning task which refers to identifying clusters of similar data samples in a data set. The K-means algorithm~\cite{Bishop} is one of the most popular hard clustering
 algorithms that has been used in many fields such as data mining and machine learning.

Despite its simplicity, running K-means on large-scale data sets presents new challenges and considerable efforts have been made to introduce memory/computation efficient clustering algorithms. In this paper,  we present a variant of the K-means algorithm which allows us to find a set of cluster centers as well as assignment of the data. The idea is to precondition and sample the data in one pass over the data to achieve a sparse matrix, therefore reducing processing time and saving memory, and applicable to streaming and distributed data. 

First, we review the standard K-means algorithm. Given a data set $\mathbf{X}=[\mathbf{x}_{1},\ldots,\mathbf{x}_{n}]\in\mathbb{R}^{p\times n}$, the goal of K-means is to partition the data into a known number of $K$ clusters such that $\boldsymbol{\mu}_{k}\in\mathbb{R}^{p}$ is the prototype associated with the $k$-th cluster for $k=1,\ldots,K$. We also introduce a set of binary indicator variables $
c_{ik}\in\{0,1\}$ to represent the assignments, where
$\mathbf{c}_{i}=[c_{i1},\ldots,c_{iK}]^{T}$ is the $k$-th canonical basis vector in $\mathbb{R}^{K}$ if and only if $\mathbf{x}_{i}$ belongs to the $k$-th cluster. 

Let us define cluster centers $\boldsymbol{\mu}=\{\boldsymbol{\mu}_{k}\}_{k=1}^{K}$ and the assignments of the data samples $\mathbf{c}=\{\mathbf{c}_{i}\}_{i=1}^{n}$. The K-means algorithm attempts to minimize the sum of the squared Euclidean distances of each data point to its assigned cluster:
\begin{equation}
J(\mathbf{c},\boldsymbol{\mu})=\sum_{i=1}^{n}\sum_{k=1}^{K}c_{ik}\left\Vert \mathbf{x}_{i}-\boldsymbol{\mu}_{k}\right\Vert _{2}^{2}.\label{K-means-obj}
\end{equation}
The objective $J(\mathbf{c},\boldsymbol{\mu})$
is minimized by 
 an iterative algorithm
 that (step one)
 updates assignments $\mathbf{c}$ and (step two)
updates  $\boldsymbol{\mu}$ as follows:\\ 
\textbf{Step 1: Minimize $J(\mathbf{c},\boldsymbol{\mu})$
    over $\mathbf{c}$, keeping $\boldsymbol{\mu}$ fixed:} 
\begin{equation}
\forall\;
\; i=1,\ldots,n:\; c_{ik}=\begin{cases}
1, &  k=\argmin_{j}\left\Vert \mathbf{x}_{i}-\boldsymbol{\mu}_{j}\right\Vert _{2}^{2}\\
0, & \text{otherwise}
\end{cases}
\end{equation}
\textbf{Step 2: Minimize $J(\mathbf{c},\boldsymbol{\mu})$
    over $\boldsymbol{\mu}$, keeping $\mathbf{c}$ fixed:}
\begin{equation}
\forall\; k=1,\ldots K:\;\boldsymbol{\mu}_{k}=\frac{1}{n_{k}}\sum_{i\in\mathcal{I}_{k}}\mathbf{x}_{i}\label{KMeansStep2}
\end{equation}
where $\mathcal{I}_{k}$ denotes the set of samples assigned to the $k$-th cluster in Step 1 and $n_{k}=\left|\mathcal{I}_{k}\right|$. Therefore, the update formula for cluster center $\boldsymbol{\mu}_{k}$ is the sample mean of the data samples in $\mathcal{I}_{k}$. To initialize K-means, a set of cluster centers can be chosen uniformly at random from the data set $\X$. However, we use the recent K-means++ algorithm~\cite{kmeans_plusplus} to choose the initial cluster centers since it improves the performance of K-means over the worst-case random initializations.

Next, we consider a probabilistic mixture model to find an optimal objective function for our sparsified K-means clustering algorithm using Maximum-Likelihood (ML) estimation.
\subsection{Optimal Objective Function via ML Estimation}
One appealing aspect of the K-means algorithm is that the objective function~\eqref{K-means-obj} coincides with the log-likelihood function of a mixture of $K$ Gaussian components (clusters), with mean $\boldsymbol{\mu}_{k}$ and covariance matrix $\boldsymbol{\Sigma}=\lambda\mathbf{I}_{p}$, where $\lambda>0$ is fixed or ``known'' (its actual value is unimportant), and treating $\mathbf{c}$ and $\boldsymbol{\mu}$ as unknown parameters.  We show that under the same assumptions, K-means clustering on sampled preconditioned data enjoys the same ML interpretation.     

Let $p(\x_i|\boldsymbol{\mu} ,\mathbf{c}_{i})$ denote the conditional probability distribution of sample $\x_i$ given that a set of centers $\boldsymbol{\mu}$ and a particular value of $\mathbf{c}_{i}$ are known. Under this setup, the conditional distribution of $\x_i$ is Gaussian and it can be written as follows:
\begin{equation}
 p\left(\mathbf{x}_{i}|\boldsymbol{\mu} ,\mathbf{c}_{i}\right)=\prod_{k=1}^{K}p\left(\mathbf{x}_{i}|\boldsymbol{\mu}_{k},\boldsymbol{\Sigma}\right)^{c_{ik}}
\end{equation}
where
 \[
	 p\left(\mathbf{x}_{i}|\boldsymbol{\mu}_{k},\boldsymbol{\Sigma}\right)=\frac{1}{\left(2\pi\lambda\right)^{\frac{p}{2}}}\exp\Big({-}\frac{1}{2\lambda}\left\Vert \mathbf{x}_{i}-\boldsymbol{\mu}_{k}\right\Vert _{2}^{2}\Big).
 \]
Given that $\x_i$ belongs to the $k$-th cluster with mean $\boldsymbol{\mu}_k$ and covariance $\boldsymbol{\Sigma}=\lambda\eye_p$, then the preconditioned data $\y_i=\Hadamard\Diag\x_i$  also has a Gaussian distribution with mean $\boldsymbol{\mu}_k':=\E[\y_i]=\Hadamard\Diag\boldsymbol{\mu}_k$ and the same covariance $\boldsymbol{\Sigma}=\lambda\eye_p$ because $\Hadamard$ and $\Diag$ are orthonormal matrices, i.e., $(\Hadamard\Diag)(\Hadamard\Diag)^{T}=\eye_p$. Note that we can also find $\boldsymbol{\mu}_k$ from $\boldsymbol{\mu}_k'$ using the equation:
\begin{equation}
\boldsymbol{\mu}_k=(\Hadamard\Diag)^{T}\boldsymbol{\mu}_k'.\label{eq:mu-mu'} 
\end{equation}
We now take $n$ independent \textit{realizations} of the sampling matrix $\mathbf{R}$, denoted $\mathbf{R}_{1},\ldots,\mathbf{R}_{n}$, each consisting of $m$ canonical basis vectors. Then, given that $\mathbf{y}_{i}$ belongs to the $k$-th cluster with mean $\boldsymbol{\mu}_{k}'$ and covariance matrix $\boldsymbol{\Sigma}=\lambda\mathbf{I}_{p}$, the sub-sampled data $\mathbf{z}_{i}=\mathbf{R}_{i}^{T}\mathbf{y}_{i}$ also has a Gaussian distribution with mean $\mathbb{E}\left[\mathbf{z}_{i}\right]=\mathbf{R}_{i}^{T}\boldsymbol{\mu}_{k}'$ and covariance $\lambda\mathbf{I}_{m}$, since $\RR_i^{T}\RR_i=\eye_m$ based on Thm.~\ref{thm:Properties_Sketching}. Hence 
\[
p\left(\mathbf{z}_{i}|\boldsymbol{\mu}_{k}',\boldsymbol{\Sigma}\right)=\frac{1}{\left(2\pi\lambda\right)^{\frac{m}{2}}}\exp\Big({-}\frac{1}{2\lambda}\left\Vert \mathbf{z}_{i}-\mathbf{R}_{i}^{T}\boldsymbol{\mu}_{k}'\right\Vert _{2}^{2}\Big)
\]
and, thus, we have the following expression for the conditional distribution of $\mathbf{z}_i$: 
\begin{equation}
p\left(\mathbf{z}_{i}|\boldsymbol{\mu}',\mathbf{c}_{i}\right)=\prod_{k=1}^{K}p\left(\mathbf{z}_{i}|\boldsymbol{\mu}_{k}',\boldsymbol{\Sigma}\right)^{c_{ik}}.
\end{equation}
Next, we consider ML estimation when we have access to the sampled preconditioned data $\mathbf{Z}=[\mathbf{z}_{1},\ldots,\mathbf{z}_{n}]$:
	\[
	p\left(\mathbf{Z}|\boldsymbol{\mu}',\mathbf{c}\right)=\prod_{i=1}^{n}\prod_{k=1}^{K}p\left(\mathbf{z}_{i}|\boldsymbol{\mu}_{k}',\boldsymbol{\Sigma}\right)^{c_{ik}}
	\]
	and taking the logarithm of the likelihood function:
	\[
	\log p\!\left(\mathbf{Z}|\boldsymbol{\mu}',\mathbf{c}\right)\hspace{-1mm}=\hspace{-1mm}{-}\frac{mn}{2}\!\log\!\left(2\pi\lambda\right)-\frac{1}{2\lambda}\!\sum_{i=1}^{n}\sum_{k=1}^{K}\hspace{-1mm}c_{ik}\left\Vert \mathbf{z}_{i}\hspace{-1mm}-\hspace{-1mm}\mathbf{R}_{i}^{T}\boldsymbol{\mu}_{k}'\right\Vert _{2}^{2}.
	\]
	Hence, the ML estimate of the unknown parameters $\mathbf{c}$ and $\boldsymbol{\mu}'$ (or equivalently $\boldsymbol{\mu}$) is obtained by minimizing:
	\begin{equation}
	J'(\mathbf{c},\boldsymbol{\mu}')=\sum_{i=1}^{n}\sum_{k=1}^{K}c_{ik}\left\Vert \mathbf{z}_{i}-\mathbf{R}_{i}^{T}\boldsymbol{\mu}_{k}'\right\Vert _{2}^{2}.\label{eq:CK_means}
	\end{equation}
	Note that $J'(\mathbf{c},\boldsymbol{\mu}')$ can be written as:
	\begin{equation}
			J'(\mathbf{c},\boldsymbol{\mu}')=\sum_{i=1}^{n}\sum_{k=1}^{K}c_{ik}\left\Vert \mathbf{R}_{i}^{T}\left(\mathbf{y}_{i}-\boldsymbol{\mu}_{k}'\right)\right\Vert _{2}^{2}\label{eq:CK_means2}
	\end{equation}
	and for $\mathbf{R}_{i}=\mathbf{I}_{p}$, $i=1,\ldots,n$, it reduces to the objective function of the standard K-means~\eqref{K-means-obj} because the preconditioning transformation $\Hadamard\Diag$ is an orthonormal matrix. 
\subsection{The Sparsified K-means Algorithm}
	Similar to the K-means algorithm, we minimize the objective function $J'(\mathbf{c},\boldsymbol{\mu}')$ in an iterative procedure that (step one) updates assignments $\mathbf{c}$ and (step two) updates $\boldsymbol{\mu}'$: \\
\textbf{Step 1: Minimize $J'(\mathbf{c},\boldsymbol{\mu}')$ holding $\boldsymbol{\mu}'$ fixed:} 	

In~\eqref{eq:CK_means}, the terms involving different $n$ are independent and we assign each sampled preconditioned data $\mathbf{z}_{i}=\RR_i^{T}\y_i\in\R^{m}$ to the closest cluster:
\begin{equation} \label{eq:assignments}
\forall\;
\; i=1,\ldots,n:\; c_{ik}=\begin{cases}
1, &  k=\argmin_{j}\left\Vert \mathbf{z}_{i}-\mathbf{R}_{i}^{T}\boldsymbol{\mu}_{j}'\right\Vert _{2}^{2}\\
0, & \text{otherwise}
\end{cases}
\end{equation}
The connection between this step and the first step of K-means is immediate mainly due to the Johnson-Lindenstrauss (JL) lemma. In fact, the accuracy of this step depends on the preservation of the Euclidean norm under selecting $m$ entries of a $p$-dimensional vector. Based on the well-known fast JL transform~\cite{FJLT}, one needs to first smooth out data samples with a few large entries to ensure the preservation of the Euclidean norm with high probability. In particular, the author in~\cite{ImprovedAnalysis} showed that selecting $m$ entries of the preconditioned data uniformly at random without replacement preserves the geometry of the data as well as the Euclidean norm. A direct consequence of this result stated in Thm.~\ref{thm:ROS-distance-preserve} shows that pairwise distances between each point and cluster centers are preserved.\\
\textbf{Step 2: Minimize $J'(\mathbf{c},\boldsymbol{\mu}')$ holding $\mathbf{c}$ fixed:}

Given the assignments $\mathbf{c}$ from Step 1, we can write~\eqref{eq:CK_means} as:
\begin{equation}
J'(\boldsymbol{\mu}')=\sum_{i\in\mathcal{I}_{k}}\left\Vert \mathbf{R}_{i}^{T}\left(\mathbf{y}_{i}-\boldsymbol{\mu}_{k}'\right)\right\Vert _{2}^{2}
\end{equation}
where $\mathcal{I}_{k}$ represents the set of samples assigned to the $k$-th cluster and recall that $n_{k}=\left|\mathcal{I}_{k}\right|$. The terms involving different $k$ are independent and each term is a quadratic function of $\boldsymbol{\mu}_{k}'$. Thus, each term can be minimized individually by setting its derivative with respect to $\boldsymbol{\mu}_{k}'$ to zero giving:
\begin{equation}
\left(\sum_{i\in\mathcal{I}_{k}}\mathbf{R}_{i}\mathbf{R}_{i}^{T}\right)\boldsymbol{\mu}_{k}'=\sum_{i\in\mathcal{I}_{k}}\mathbf{R}_{i}\mathbf{R}_{i}^{T}\mathbf{y}_{i}.\label{eq:mu_update_one}
\end{equation}
Note that $\sum_{i\in\mathcal{I}_{k}}\mathbf{R}_{i}\mathbf{R}_{i}^{T}\in \mathbb{R}^{p\times p}$ is a diagonal matrix, where its $j$-th diagonal element counts the number of cases the $j$-th canonical basis vector is chosen in the sampling matrices $\mathbf{R}_{i}$ for all $i\in\mathcal{I}_{k}$, denoted by $n_{k}^{(j)}$. Therefore, for any $j$ with $n_{k}^{(j)}=0$, we cannot estimate the $j$-th entry of $\boldsymbol{\mu}_{k}'$ and the corresponding entry should be removed from~\eqref{eq:mu_update_one}. Given that $n_{k}^{(j)}>0$ for all $j$, $\boldsymbol{\mu}_{k}'$ is updated as follows: 
\begin{equation}
	\boldsymbol{\mu}_{k}'=\text{diag}\left(\Big[\frac{1}{n_{k}^{(1)}},\ldots,\frac{1}{n_{k}^{(p)}}\Big]\right)\left(\sum_{i\in\mathcal{I}_{k}}\mathbf{R}_{i}\mathbf{R}_{i}^{T}\mathbf{y}_{i}\right).\label{eq:mu_update_two}
\end{equation}
Recall that each $\mathbf{R}_{i}\mathbf{R}_{i}^{T}\mathbf{y}_{i}$ is the sampled preconditioned data such that $m$ out $p$ entries are kept uniformly at random. Hence, the update formula for $\boldsymbol{\mu}_{k}'$ is the \textit{entry-wise sample mean} of the sparse data samples in the $k$-th cluster. The sparsified K-means algorithm is summarized in Algorithm~\ref{alg:sparsified-kmeans}. 
 \begin{algorithm}
 	\textbf{Input:} Dataset $\mathbf{X}\in\mathbb{R}^{p\times n}$, number
 	of clusters $K$, compression factor $\gamma=\frac{m}{p}<1$. 
 	
 	\textbf{Output:} Assignments $\mathbf{c}=\{\mathbf{c}_{i}\}_{i=1}^{n}\in\mathbb{R}^{K}$, cluster centers $\boldsymbol{\mu}=\{\boldsymbol{\mu}_k\}_{k=1}^{K}\in\mathbb{R}^{p}$. 
 	\caption{Sparsified K-means}	\label{alg:sparsified-kmeans}
 	\begin{algorithmic}[1]
 		\Function{Sparsified K-means}{$\X,K,\gamma$}
 		\State $\mathbf{Y}\leftarrow\mathbf{H}\mathbf{D}\mathbf{X}$ \Comment{See Eq.~\eqref{eq:ROS} }
 		\For{$i=1,\ldots,n$}
 		\State $\mathbf{w}_{i}=\mathbf{R}_{i}\mathbf{R}_{i}^{T}\mathbf{y}_{i}$  \Comment{$\RR_i\in\R^{p\times m}$:sampling matrix}
 		\EndFor
          		\State Find initial cluster centers via K-means++~\cite{kmeans_plusplus} 
 		\For{each iteration}
 		\State update assignments $\mathbf{c}$ \Comment{See Eq.~\eqref{eq:assignments} }
 		\State update cluster centers $\boldsymbol{\mu}'$ \Comment{See Eq.~\eqref{eq:mu_update_two} } 
 		\EndFor 
 		\State $\boldsymbol{\mu}=\left(\Hadamard\Diag\right)^{T}\boldsymbol{\mu}'$ \Comment{See Eq.~\eqref{eq:mu-mu'}}
 		\State Return $\mathbf{c}$ and $\boldsymbol{\mu}$. 
 		\EndFunction
 	\end{algorithmic}
 \end{algorithm}

\begin{figure}
		\centering
		\includegraphics[width=\figWidth]{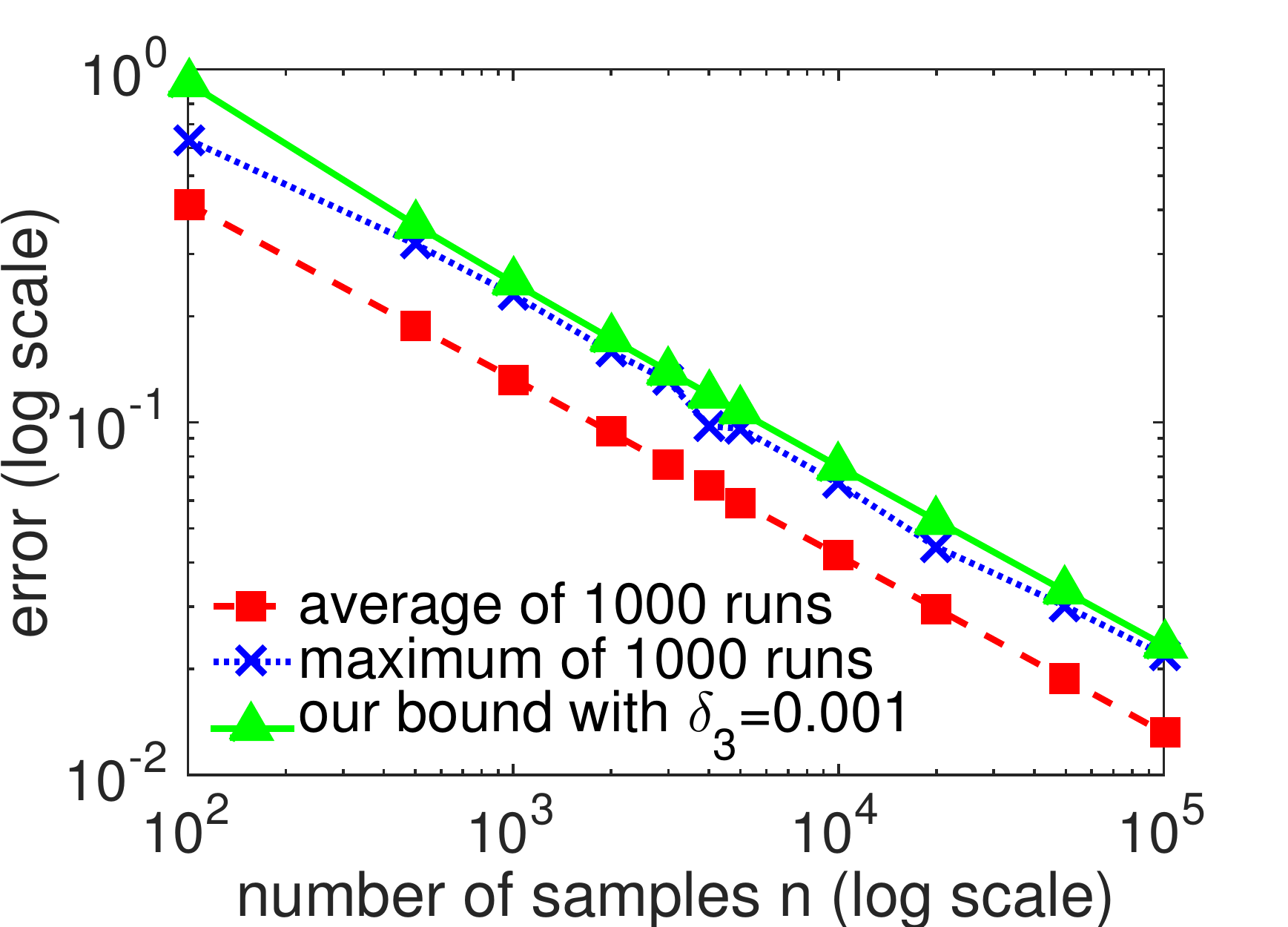}
		\caption{Verifying the accuracy of Thm.~\ref{thm:RRT}. We set parameters $p=100$ and $\gamma=m/p=0.3$ and plot the average and maximum of $\left\Vert \mathbf{H}_{k}-\mathbf{I}_{p}\right\Vert_2$ over $1000$ runs for varying $n$. We compare the empirical values with our theoretical bound when $\delta_3=0.001$. We see that our bound is tight.  
			\label{fig:convergence_sum_RRT}
		}
	\end{figure}

Now, we return to equation~\eqref{eq:mu_update_one} and study the accuracy of the estimated solution $\boldsymbol{\mu}_{k}'$. To do this, we re-write~\eqref{eq:mu_update_one} as:
\begin{equation}
\mathbf{H}_{k}\boldsymbol{\mu}_{k}'=\mathbf{m}_{k}\label{eq:mu_k_update}
\end{equation}
where 
\begin{equation}
\mathbf{H}_{k}=\frac{p}{m}\frac{1}{n_{k}}\sum_{i\in\mathcal{I}_{k}}\mathbf{R}_{i}\mathbf{R}_{i}^{T},\;\mathbf{m}_{k}=\frac{p}{m}\frac{1}{n_{k}}\sum_{i\in\mathcal{I}_{k}}\mathbf{R}_{i}\mathbf{R}_{i}^{T}\mathbf{y}_{i}.\label{Hkmk}
\end{equation}
Next, we show that $\mathbf{H}_{k}$ converges to the identity matrix $\mathbf{I}_{p}$ as the number of samples in each cluster $n_k$ increases.  
\begin{thm}\label{thm:RRT}
	Consider $\mathbf{H}_{k}$ defined in~\eqref{Hkmk}. Then, for all $t\geq0$:
	\begin{equation}
	\mathbb{P}\left\{ \left\Vert \mathbf{H}_{k}-\mathbf{I}_{p}\right\Vert _{2}\leq t\right\} \geq 1-\delta_{3}
	\end{equation}
	where the failure probability,
	\begin{equation}
	 \delta_{3}=p\exp\left(\frac{-n_{k}\nicefrac{t^{2}}{2}}{\left(\frac{p}{m}-1\right)+\left(\frac{p}{m}+1\right)\nicefrac{t}{3}}\right).\label{eq:delta3}
	\end{equation}
\end{thm}
\begin{IEEEproof}
	We can write $\mathbf{S}=\mathbf{H}_{k}-\mathbf{I}_{p}=\sum_{i=1}^{n_{k}}$$\mathbf{Z}_{i}$,
	where
	
	\[
	\mathbf{Z}_{i}=\frac{1}{n_{k}}\left(\frac{p}{m}\mathbf{R}_{i}\mathbf{R}_{i}^{T}-\mathbf{I}_{p}\right),\; i=1,\ldots,n_{k}
	\]
	are independent and symmetric random matrices. Moreover, we have
	$\mathbb{E}[\mathbf{Z}_{i}]=\mathbf{0}$ using Thm.~\ref{thm:Properties_Sketching}. To apply
	the matrix Bernstein inequality given in Appendix~\ref{sec:concentration_ineq}, we should find a uniform bound on the spectral norm of each summand $\|\Z_i\|_2$:
	\begin{equation}
	\left\Vert \mathbf{Z}_{i}\right\Vert _{2}\leq\frac{1}{{n_{k}}}\left(\left\Vert \frac{p}{m}\mathbf{R}_{i}\mathbf{R}_{i}^{T}\right\Vert _{2}+\left\Vert \mathbf{I}_{p}\right\Vert _{2}\right)=\frac{1}{n_{k}}\left(\frac{p}{m}+1\right)\label{eq:bound_Zi_Hk}
	\end{equation}
	where it follows from the triangle inequality for the spectral norm and the
	fact that $\mathbf{R}_{i}\mathbf{R}_{i}^{T}\in\mathbb{R}^{p\times p}$
	is a diagonal matrix with only $m$ ones on the diagonal and the rest
	equal to zero. 
	
	Next, we find $\mathbb{E}[\mathbf{Z}_{i}^{2}]$ using the results
	of Thm.~\ref{thm:Properties_Sketching}:
	\[
	\mathbb{E}\left[\mathbf{Z}_{i}^{2}\right]\hspace{-1mm}=\hspace{-1mm}\frac{1}{n_{k}^{2}}\mathbb{E}\Big[\Big(\frac{p^{2}}{m^{2}}-2\frac{p}{m}\Big)\mathbf{R}_{i}\mathbf{R}_{i}^{T}+\mathbf{I}_{p}\Big]\hspace{-1mm}=\hspace{-1mm}\frac{1}{n_{k}^{2}}\Big(\frac{p}{m}-1\Big)\mathbf{I}_{p}
	\]
	and thus 
	\begin{equation}
	\sigma^{2}=\Big\Vert \sum_{i=1}^{n_{k}}\mathbb{E}\left[\mathbf{Z}_{i}^{2}\right]\Big\Vert _{2}=\frac{1}{n_{k}}\left(\frac{p}{m}-1\right).\label{eq:Variance_Zi_Hk}
	\end{equation}
	We now use Theorem~\ref{thm:MatrixBernstein} and this completes the proof. 
	\end{IEEEproof}
	To verify the accuracy of Thm.~\ref{thm:RRT}, we consider a numerical experiment. We set the parameters $p=100$ and compression factor $\gamma=m/p=0.3$ and show the closeness of $\mathbf{H}_k$ to $\eye_p$ for various values of $n$ over $1000$ runs. For each value of $n$, we generate $n$ sampling matrices $\RR_i$ consisting of $m$ distinct canonical basis vectors uniformly at random. We report the average and maximum of empirical values $\|\mathbf{H}_k-\eye_p\|_2$ over $1000$ runs in Fig.~\ref{fig:convergence_sum_RRT}. We also compare the empirical values with our theoretical error bound $t$ in~\eqref{eq:delta3}, when the failure probability $\delta_3=0.001$. We see that our error bound is tight and very close to the maximum of $1000$ since $\delta_3=0.001$. 
	
	Now, we present a theorem to show the connection between the updated cluster center in our sparsified K-means in~\eqref{eq:mu_k_update} and the update formula for the standard K-means algorithm.

	\begin{thm}\label{thm:accuracy-sparse-k-means}
	 Consider the update formula for the $k$-th cluster center in our sparsified K-means algorithm $\boldsymbol{\mu}_k=(\Hadamard\Diag)^{T}\boldsymbol{\mu}_k'$, where $\boldsymbol{\mu}_k'$ is given by the equation $\mathbf{H}_k\boldsymbol{\mu}_k'=\mathbf{m}_k$ in~\eqref{eq:mu_k_update}. Let $\overline{\x}_k$ denote the sample mean of the data samples in the $k$-th cluster, i.e., $\overline{\x}_k=\frac{1}{n_{k}}\sum_{i\in\mathcal{I}_{k}}\mathbf{x}_{i}$, which is the standard update formula for K-means on $\X$. Then, for all $t\geq0$,
	 \begin{equation}
	 \frac{1}{\sqrt{p}}\|\boldsymbol{\mu}_k-\overline{\x}_k\|_2\leq t\left(1+\frac{1}{\sqrt{p}}\|\boldsymbol{\mu}_k\|_2\right)
	 \end{equation}
	 with probability greater than $1-\max\{\delta_1,\delta_3\}$, where 
	 \begin{equation}
	 \delta_{1}=2p\exp\left(\frac{-n_{k}t^{2}/2}{(\frac{p}{m}-1)\nicefrac{\|\Y\|_\maxRow^{2}}{n}+\nicefrac{\tau(m,p)\left\Vert \mathbf{Y}\right\Vert_{max}t}{3}    }\right),
	 \end{equation}
	 $\tau(m,p)$ is defined in~\eqref{eq:tau-m-p} and $\delta_3$ is given in~\eqref{eq:delta3}. Recall that $\Y=\Hadamard\Diag\X$ is the preconditioned data in our sparsified K-means algorithm.
	 \end{thm}
	 \begin{IEEEproof}
	 Based on Thm.~\ref{thm:sampleMean} and Thm.~\ref{thm:RRT}, we re-write equation~\eqref{eq:mu_k_update} as:
	 \[
	 \left(\eye_p+\mathbf{E}\right)\boldsymbol{\mu}_k'= \Hadamard\Diag\overline{\x}_k + \mathbf{e}
	 \]
	 where $\|\mathbf{E}\|_2\leq t$ and $\|\mathbf{e}\|_2 \leq \sqrt{p}t$ with probabilities greater than $1-\delta_3$ and $1-\delta_1$ respectively. Thus, with probability greater than $1-\max\{\delta_1,\delta_3\}$, we have:
	 \begin{eqnarray}
	 &&\hspace{-15mm}\|\boldsymbol{\mu}_k'-\Hadamard\Diag\overline{\x}_k\|_2=\|\mathbf{e}-\mathbf{E}\boldsymbol{\mu}_k'\|_2 \nonumber\\
	 &&\hspace{-15mm}\leq \|\mathbf{e}\|_2 + \|\mathbf{E}\boldsymbol{\mu}_k'\|_2\leq \|\mathbf{e}\|_2+\|\mathbf{E}\|_2\|\boldsymbol{\mu}_k'\|_2 \nonumber\\
	 &&\hspace{-15mm}\leq t\sqrt{p}\left(1+\frac{\|\boldsymbol{\mu}_k'\|_2}{\sqrt{p}}\right)
	 \end{eqnarray}
	 where we used the triangle inequality for the spectral norm. Recall that $\Hadamard\Diag$ is an orthonormal matrix and $\boldsymbol{\mu}_k'=\Hadamard\Diag\boldsymbol{\mu}_k$. Thus, $\|\boldsymbol{\mu}_k'-\Hadamard\Diag\overline{\x}_k\|_2=\|\boldsymbol{\mu}_k-\overline{\x}_k\|_2$ and $\|\boldsymbol{\mu}_k'\|_2=\|\boldsymbol{\mu}_k\|_2$ and this completes the proof. 
	 \end{IEEEproof}

\section{Numerical Experiments}\label{sec:numerical_experiments}

We implement the sparsified K-means algorithm in a mixture of Matlab and C, available online\footnote{\url{https://github.com/stephenbeckr/SparsifiedKMeans}}. Since K-means attempts to minimize a non-convex objective, the starting points have a large effect. We use the recent K-means++ algorithm~\cite{kmeans_plusplus} for choosing starting points, and re-run the algorithm for different sets of starting points and then choose the results with the smallest objective value. All results except the big-data tests use $20$ different starting trials.

Timing results are from running the algorithm on a desktop computer with two Intel Xeon EF-2650 v3 CPUs at $2.4$--$3.2$ GHz and 8 cores and $20$~MB cache each, and should be interpreted carefully. First, we note that K-means is iterative and so the number of iterations may change slightly depending on the variant. 
Furthermore, none of the code was optimized for small problems, so timing results under about $10$ seconds do not scale with $n$ and $p$ as they do at large scale. At the other extreme, 
our first series of tests are not on out-of-core data, so the benefits of a single-pass algorithm are not apparent. Our subsequent tests are out-of-core implementations, meaning that they explicitly load data from the hard drive to RAM as few times as possible, and so the number of passes through the data becomes relevant, cf.\ Table~\ref{table:passes}.

We also caution about interpreting the accuracy results for correct identification of clusters. In our experience, if the accuracy is greater than about $75\%$, then using the result as the initial value for a single-step of standard K-means, thus increasing the number of passes through the data by one, is sufficient to match the accuracy of the standard K-means algorithm.

\begin{table}
    \caption{Low-pass Algorithms for K-means clustering \label{table:passes}}
    \centering
    \begin{tabular}{lcc}
        \toprule
          & \multicolumn{2}{c}{Passes through data...} \\
        Algorithm & ...to find $\boldsymbol{\mu}$ & ...to find $\mathbf{c}$ \\
        \midrule
        Sparsified K-means (1-pass) & 1 & 1 \\
        Sparsified K-means (2-pass) & 2 & 2 \\
        Feature extraction & 2 & 1 \\
        Feature selection & 4 & 3 \\
        \bottomrule
    \end{tabular} 
\end{table}

A two-pass sparsified K-means algorithm can be constructed by running the one-pass sparsified K-means in Algorithm~\ref{alg:sparsified-kmeans} to compute the assignments as well as the cluster centers, then re-computing the cluster center estimates $\boldsymbol{\mu}$ as the average of their assigned data points in the original (non-sampled) domain. Meanwhile, we can re-assign the data samples to the cluster centers in the original domain based on the previous center estimates from one-pass sparsified K-means. The same extra-pass procedure \emph{must} be applied to feature extraction and feature selection, since their default center estimates $\boldsymbol{\mu}$ are in a compressed domain.

\begin{algorithm}
\caption{Sparsified K-means, 2-pass}
\label{algo:kmeans2}
\begin{algorithmic}[1]
\Function{Sparsified K-means 2-pass}{$\X,K,\gamma$}
    \State  $(\mathbf{\hat{c}},\boldsymbol{\hat{\mu}})=$\Call{Sparsified K-means}{$\X,K,\gamma$}\Comment{ Alg.~\ref{alg:sparsified-kmeans} }
    \For{ $k=1,\ldots,K$ }
      \State $\boldsymbol{\mu}_k=\mathbf{0},\; \mathcal{I}_{k}=\emptyset$
    \EndFor
    \For{ $i=1,\ldots,n$ }
        \State Find cluster assignment, i.e., $k$ s.t. $\mathbf{\hat{c}}_i=\mathbf{e}_k$
        \State $\boldsymbol{\mu}_{k} \mathrel{+}= \xi$,
             $\mathcal{I}_{k}=\mathcal{I}_{k}\cup\{i\}$
        \State $\mathbf{c}_i=\argmin_{k=1,\ldots,K}\|\x_i-\boldsymbol{\hat{\mu}}_k\|_2^{2}$
    \EndFor
    \For{ $k=1,\ldots,K$ } 
    \State $\boldsymbol{\mu}_{k}\leftarrow 
    \boldsymbol{\mu}_{k}/|\mathcal{I}_{k}|$
    \EndFor
   	\State Return $\mathbf{c}$ and $\boldsymbol{\mu}$. 
 
\EndFunction
\end{algorithmic}
\end{algorithm}

Our tests compare with the feature selection and feature extraction algorithms of~\cite{Randomized_Dim_K_means}. In \emph{feature selection}, one first uses  a fast approximate SVD algorithm, e.g., \cite{Martinson_SVD,boutsidis2014near}, to compute an estimate of the left singular vectors of the data matrix $\X$. Then, the selection of $m$ rows of $\X$ is done with a randomized
sampling approach with probabilities that
are computed from the estimated left singular vectors. This can be written as $\rand\X$, where rows of the sampling matrix $\rand\in\R^{m\times p}$ are chosen from the set of canonical basis vectors in $\R^p$ based on the computed probabilities. In \emph{feature extraction}, the samples are again $\rand\X$ but the sampling matrix $\rand\in\R^{m\times p}$ is set to be a random sign matrix. Thus, the computational cost of feature extraction is dominated by the matrix-matrix multiplication $\rand\X$, whereas the dominant cost in feature selection is the approximate SVD.

\subsection{Sketched K-means for faster computation}
Sampling the data leads to both computational time and memory benefits, with computational time benefits becoming more significant for more complicated algorithms, such as the Expectation-Maximization algorithm in Gaussian mixture models that require eigenvalue decompositions. Even for the standard K-means algorithm, sub-sampling leads to faster computation. The most expensive computation in K-means is finding the nearest cluster center for each point, since a naive implementation of this costs $\order(pnK)$ flops per iteration\footnote{
We do not consider the variants of K-means based on kd-trees since these have running time exponential in $p$ and are suitable for $p \lesssim 20$~
\cite{KmeansKDtree}.}.   
By effectively reducing the dimension from $p$ to $m$, the sparse version sees a speedup of roughly $\gamma^{-1}=p/m$.

\begin{figure}
\centering
\includegraphics[width=\figWidth,trim= 3 0 0 0,clip]{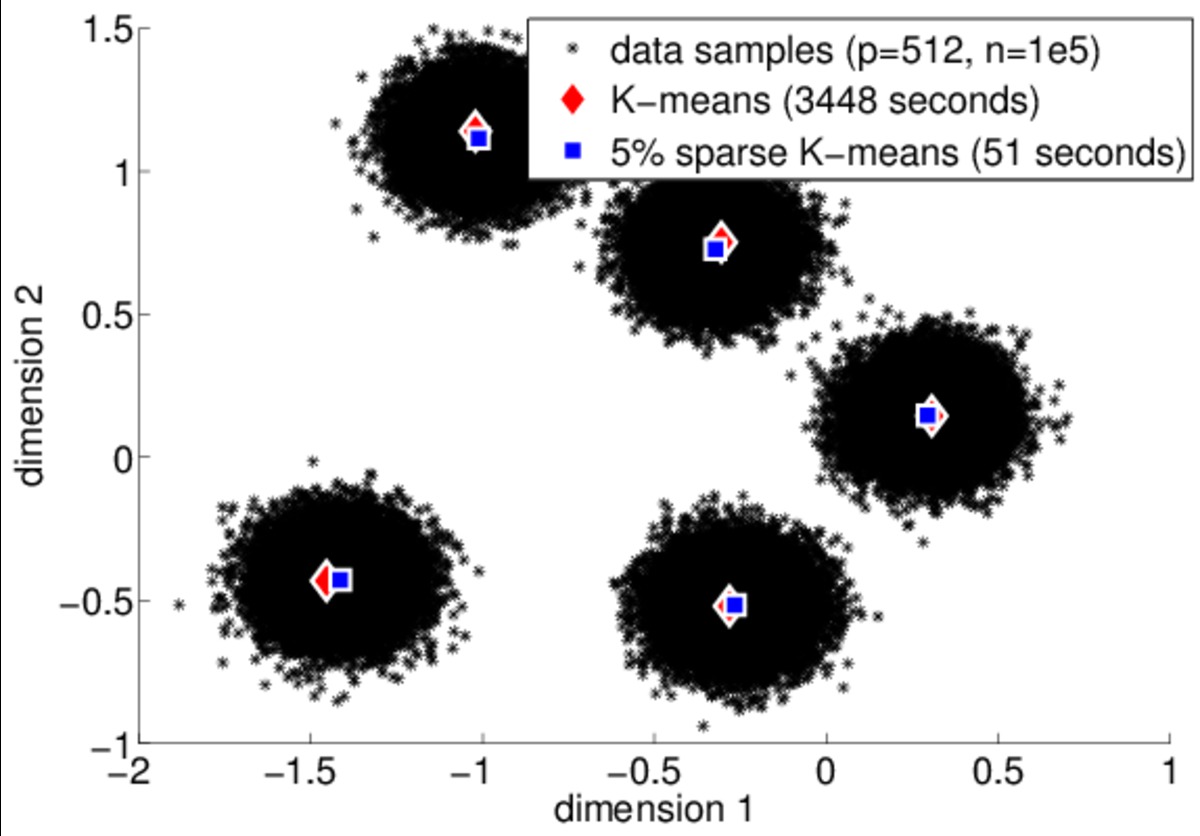} 
\caption{Standard K-means and our sparse version of K-means, on synthetic data, $n=10^5$.
    \label{fig:speedup}
}
\end{figure}

Fig.~\ref{fig:speedup} demonstrates this speedup experimentally on a toy problem. The data of size $p=512$ and $n=10^{5}$ are generated artificially so that each point belongs to one of $K=5$ clusters and is perturbed by a small amount of Gaussian noise.  An optimized variant of Matlab's \texttt{kmeans} algorithm takes $3448$ seconds to run. 

We compare this with random Hadamard mixing followed by $5\%$ sub-sampling, which takes $51$ seconds. 
 The first two dimensions of the data are shown in Fig.~\ref{fig:speedup} which makes it clear that there is no appreciable difference in quality, while our sparsified K-means algorithm is approximately $67$ times faster.

\subsection{Comparison with dimensionality-reducing approaches on real data}
For a realistic clustering application, experiments are performed on the MNIST dataset\footnote{\url{http://yann.lecun.com/exdb/mnist/}} which consists of centered versions of hand-written digits, each digit stored as a $28\times 28$ pixel image. For processing, the images are vectorized so $p=28^2=784$. 
The dataset includes both testing and training sets, though for our purposes we combined the two and report in-sample error, since the effect of sampling and dimensionality reduction to out-of-sample error is beyond our scope.  

For simplicity of interpreting results, we use data from the samples of three digits (``0'', ``3'' and ``9''), so $K=3$ in the clustering algorithm. There were $6903$, $7141$ and $6958$ examples of each class of image, respectively, so $n=21002$. 
The original data provides a ground-truth label, against which we compute accuracy by computing the total number of correctly assigned images, normalized by the total number of images. All the algorithms, except standard K-means, are stochastic, so we re-run the clustering 50 times and record the mean and standard deviation of these 50 trials. Recall that within each run, we choose the best of 20 random starting points.

\begin{figure}
    \centering
    \includegraphics[width=\figWidth]{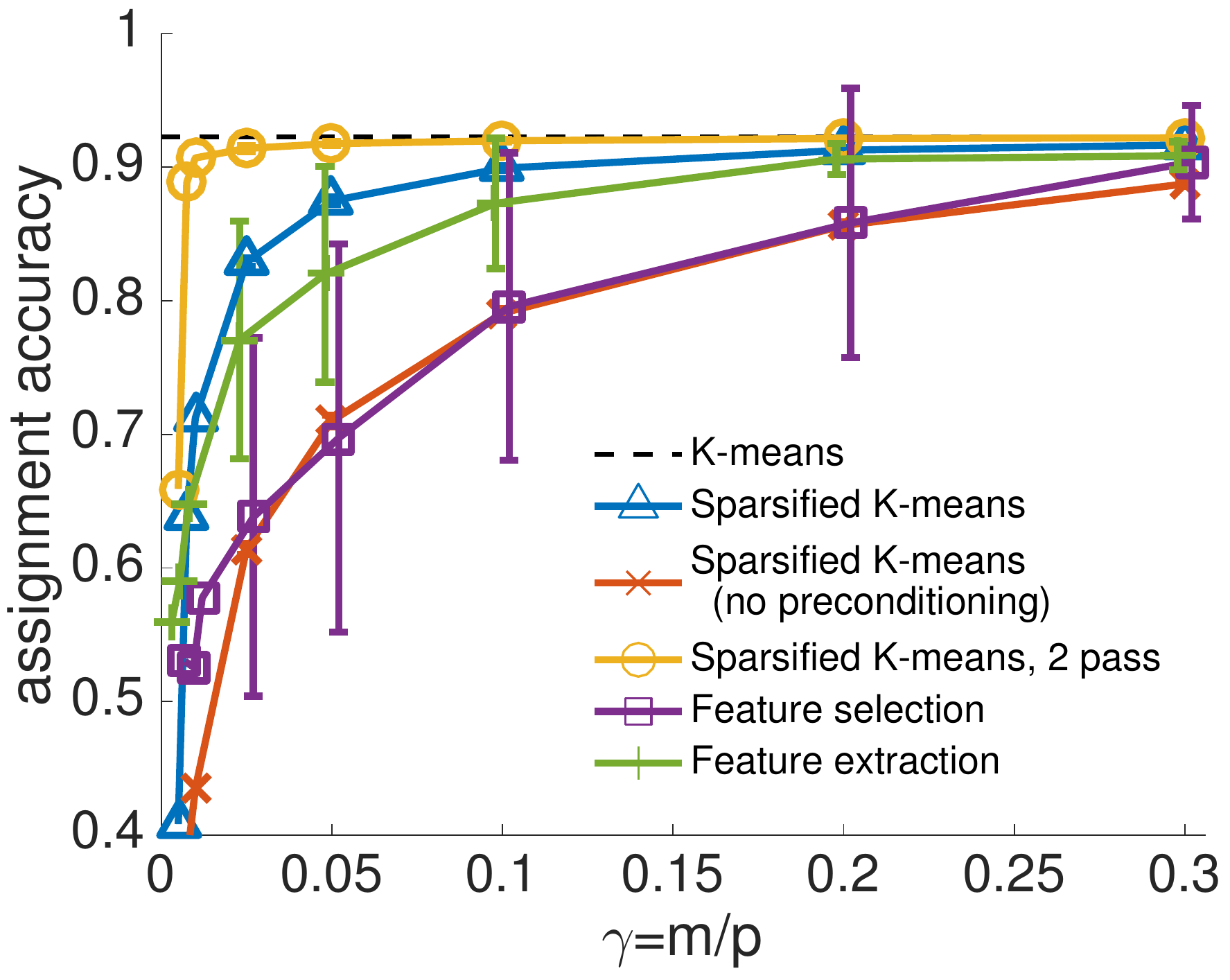}
    \caption{
        Accuracy of various K-means algorithms on the MNIST data, 50 trials. 
        The plot shows the mean and the standard deviation error bars, empirically suggesting that feature-based algorithms show higher variance than the sampling-based algorithm.
        For example, at $\gamma=0.1$, the standard deviation for sparsified K-means, sparsified K-means without preconditioning, 2-pass sparsified K-means, feature selection, and feature extraction, is
         $0.002$, $0.004$, $0.001$, $0.1151$ and $0.049$, 
         respectively. Moreover, the accuracy of 2-pass sparsified K-means reaches the accuracy of standard K-means even for small values of the compression factor $\gamma$. For visual clarity, we did not include the standard deviation error bars for $\gamma<0.025$. 
        \label{fig:MNIST:accuracy}
    }
\end{figure}

\paragraph{Timing and accuracy}

\begin{figure}
    \centering 
 \includegraphics[width=\figWidth]{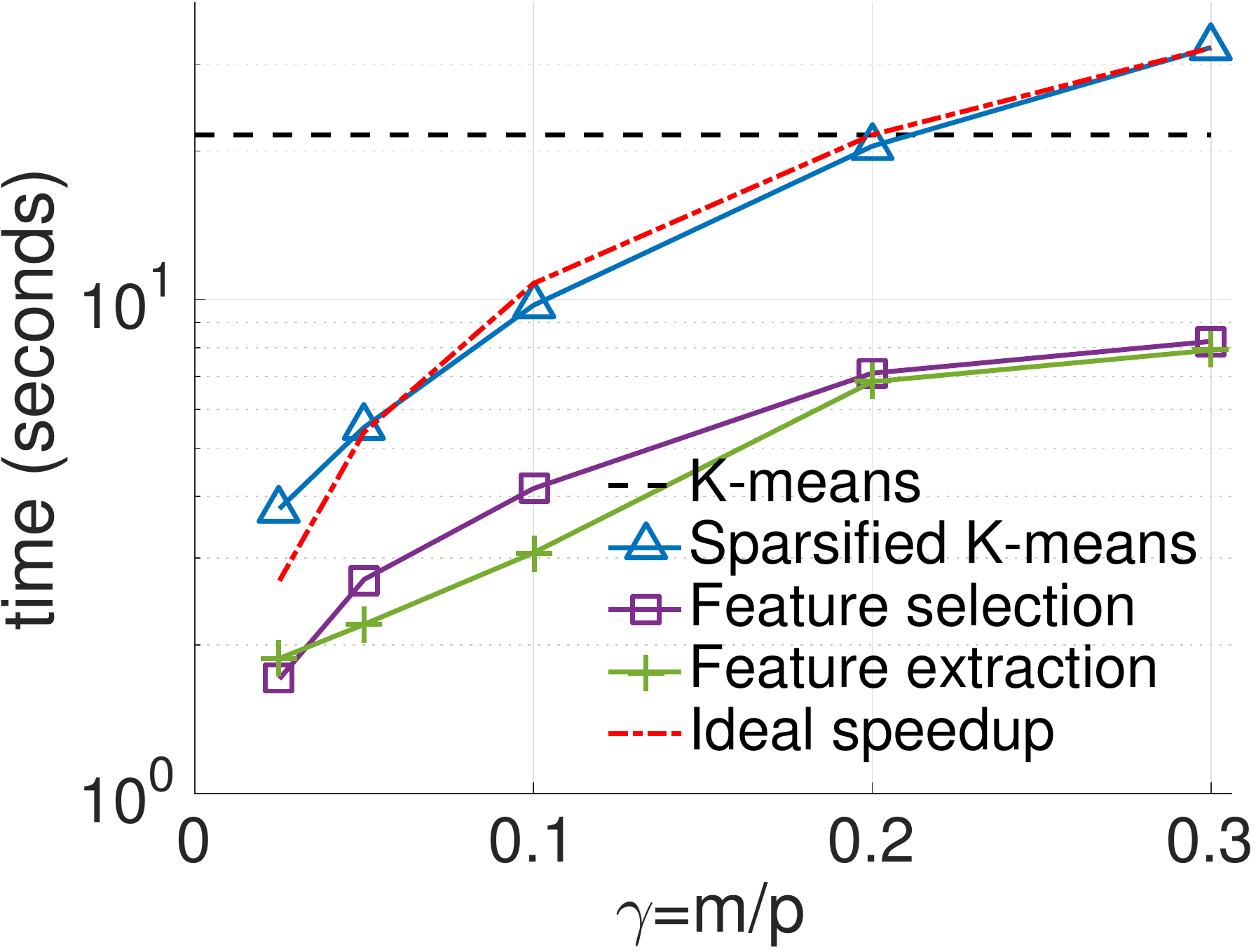}
    \caption{Timing of various K-means algorithms on the MNIST data. The three variants of sparsified K-means (with and without preconditioning, and 2-pass) all take approximately the same time on this dataset, so we only show the time for preconditioned sparsified K-means. 
The red dashed curve is proportional to $\gamma$, which is the ideal speedup ratio; the constant of proportionality is $5$, chosen to make the curve line up with the sparsified K-means performance. Note that time is in log-scale; at $\gamma=0.3$, sparsified K-means takes about 12 seconds while feature extraction takes 8 seconds.
        \label{fig:MNIST:time}
    }
\end{figure}
Clustering accuracy as a function of $\gamma$ is shown in Fig.~\ref{fig:MNIST:accuracy}, which suggests sparsified K-means is the most accurate of the efficient methods, and that accuracy is further improved in the two-pass version. 
Timing results are shown in Fig.~\ref{fig:MNIST:time}, which also shows the time for our optimized implementation of K-means on the full data.
All the efficient algorithms show a speedup over K-means proportional to $\gamma^{-1}$, as expected, until the sparsity is near $5\%$, at which point various fixed costs in the computation start to dominate; one would expect ideal speedup to continue to lower $\gamma$ if $n$ were larger, cf.~Table~\ref{table:timingBigger2}.
Sparsified K-means takes longer than standard K-means as $\gamma\rightarrow 1$ since it is inefficient to work with a sparse matrix format when the matrices are not actually sparse.

\begin{figure}
    \centering
    \subfloat[true cluster centers]{
        \includegraphics[width=.22\textwidth]{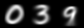}
        \label{fig:trueCenter}
    }\hspace{4mm}
    \subfloat[K-means, many passes]{
        \includegraphics[width=.22\textwidth]{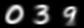}
        \label{fig:KmeansCenter}
    }

\subfloat[sparsified K-means, 1 pass, no preconditioning]{
          \includegraphics[width=.22\textwidth]{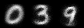}
          \label{fig:sparse_1no}
} \hspace{4mm}
\subfloat[sparsified K-means, 2 passes, no preconditioning]{
          \includegraphics[width=.22\textwidth]{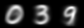}
          \label{fig:sparse_2no}
}

\subfloat[sparsified K-means, 1 pass, preconditioned]{
          \includegraphics[width=.22\textwidth]{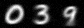}
          \label{fig:sparse_1}
}\hspace{4mm}
\subfloat[sparsified K-means, 2 passes, preconditioned]{
          \includegraphics[width=.22\textwidth]{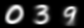}  
          \label{fig:sparse_2}
}

\subfloat[feature extraction, 1 pass]{
        \includegraphics[width=.22\textwidth]{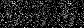}
        \label{fig:extract_1}
}\hspace{4mm}
\subfloat[feature extraction, 2 passes]{
        \includegraphics[width=.22\textwidth]{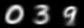}
        \label{fig:extract_2}
}
    
\subfloat[feature selection, 3 passes]{
        \includegraphics[width=.22\textwidth]{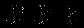}
        \label{fig:selection_3}
}\hspace{4mm}
\subfloat[feature selection, 4 passes]{
        \includegraphics[width=.22\textwidth]{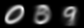}
        \label{fig:selection_4}
}

    \caption{Center estimates $\boldsymbol{\mu}$ from
        low-pass K-means algorithms.
        \label{fig:centerEstimate}
    }
\end{figure}

\paragraph{Center estimation}\label{paragraph:center}
The estimated cluster centers $\boldsymbol{\mu}$ from several low-pass K-means algorithms are shown in Fig.~\ref{fig:centerEstimate}, for $\gamma=0.03$. As we see, our sparsified K-means algorithm returns fairly accurate estimates of the true cluster centers in one pass over the data, which represent the three classes of digits in the given unlabeled dataset. However, as described, feature-based algorithms require one more pass over the full dataset after finding assignments to return meaningful estimates of the true cluster centers.

Why does our method give effective 1-pass center estimates, while the other methods do not, even if they have comparable accuracy in estimating assignments? 
The reason is that each sample $\mathbf{x}_i$ is sampled with an \emph{independent copy} of the random sampling operator $\mathbf{R}_{i}\mathbf{R}_{i}^{T}$, and this leads to a \emph{consistent} estimator. For simplicity, assume assignments have been made correctly for a given cluster $k$, so we know $\mathcal{I}_k$. 
Then Thm.~\ref{thm:RRT} bounds $\|\mathbf{H}_{k}-\mathbf{I}_{p}\|_2$ in terms of $n_k:= |\mathcal{I}_{k}|$, and in particular, we know $\mathbf{H}_{k}$ converges to $\mathbf{I}_{p}$ almost surely as $n_k \rightarrow \infty$ (this follows from the strong law of large numbers).
To be specific, recall that $\boldsymbol{\mu}_k=\frac{1}{n_k}\sum_{i\in \mathcal{I}_{k}}\xi$ (assume $\xi$ are deterministic, though the argument adapts to stochastic $\xi$ under mild assumptions such as finite first two moments), then from the sampled data we can form the center estimate
\[
 \widehat{\boldsymbol{\mu}_k} = \frac{1}{n_k} \frac{p}{m} \sum_{i\in \mathcal{I}_{k}} \mathbf{R}_{i}\mathbf{R}_{i}^{T}\x_i
\]
and $\widehat{\boldsymbol{\mu}_k}\rightarrow \boldsymbol{\mu}_k$ almost surely as $n_k\rightarrow \infty$, which follows from the strong law of large numbers and the independence of the $\mathbf{R}_{i}$.

For feature extraction (FE), the collected data are $\{ \rand\x_i \}_{i\in \mathcal{I}_{k}}$, so the obvious center estimate is \[
\widehat{\boldsymbol{\mu}_k}^\text{FE} = \frac{1}{n_k} \sum_{i\in \mathcal{I}_{k}}  \rand^\dagger \rand\x_i
= \frac{1}{n_k}  \rand^\dagger \rand \sum_{i\in \mathcal{I}_{k}} \x_i
\]
with $\dagger$ representing the pseudo-inverse. 
As $n_k\rightarrow \infty$, this does \emph{not} converge to $\boldsymbol{\mu}_{k}$ because $\boldsymbol{\Omega}^\dagger \boldsymbol{\Omega} \neq \mathbf{I}_{p}$ (equality is impossible because the term on the left has rank $m<p$). That is, even with more data, the center estimate does not improve because a single copy of the random variable $\boldsymbol{\Omega}$ is used to compress all the data, so it is not consistent. The only solution is to take another pass through the original data $\{\xi\}$ using the estimated cluster assignments to form the sample center estimate.

\subsection{Big data tests}
We further test on two increasingly large extensions of MNIST.
On the largest extension, we test an out-of-core memory version of our algorithm. Because of the size of the data, we no longer run the K-means algorithm on all the data in order to provided a benchmark, but since the data are generated similarly to the classic MNIST, it would be reasonable to expect that basic K-means would behave similarly and achieve an accuracy close to $92\%$ as in Fig.~\ref{fig:MNIST:accuracy}. 
    
Since the algorithms take longer to run, we reduce the number of replicates in each trial to $10$, and perform only 10 trials per algorithm. We focus on feature extraction and not feature selection since the smaller-scale tests indicated feature extraction performed better in both speed, accuracy and number of passes through the data.

\paragraph{In-core memory with $n=6\cdot 10^5$} 
Clustering was performed on data from the first $200,\!000$ samples each of the ``0'', ``3'' and ``9'' digits from the mnist8M dataset~\cite{loosli-canu-bottou-2006}, using code from the ``Infinite MNIST'' project\footnote{\url{http://leon.bottou.org/projects/infimnist}}, thus $n=6\cdot 10^5$, with $p=784$ as before. This new dataset artificially creates more training examples by applying pseudo-random deformations and translations to the MNIST images.

\begin{figure}
    \centering
\includegraphics[width=\figWidth]{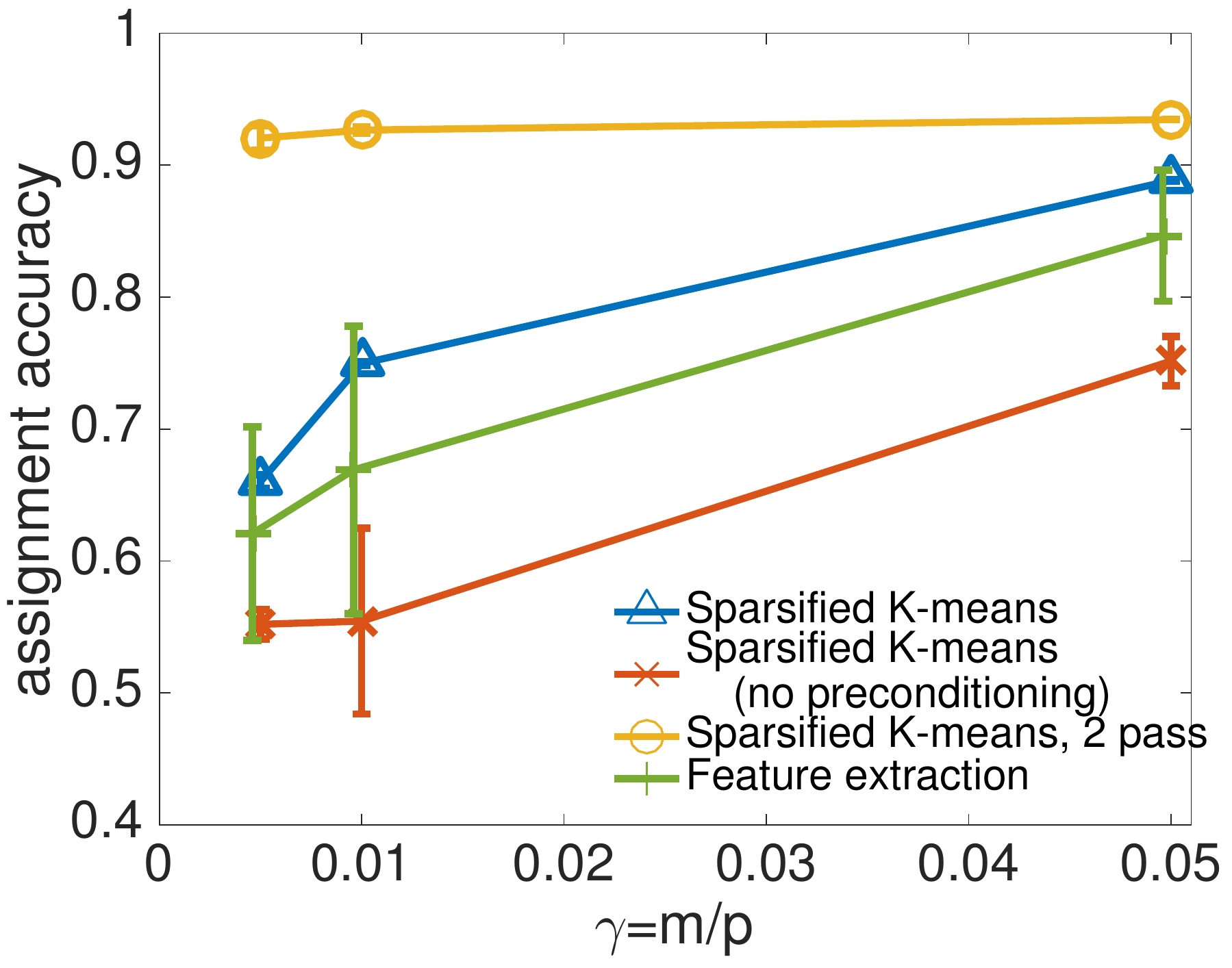}
\caption{Similar to Fig.~\ref{fig:MNIST:accuracy} but on much larger data, $n=6\cdot 10^5$. The preconditioning helps significantly to increase accuracy, as well as lower variance. }
\label{fig:mnist8mError}
\end{figure}

Accuracy results are in Fig.~\ref{fig:mnist8mError}. The preconditioned version of sparsified K-means is much more accurate than the non-preconditioned version, and has better accuracy than  feature extraction while also enjoying lower variance and taking only a single pass through the data. If we compute a second pass through the data, accuracy jumps to nearly optimal levels as soon as we sample at least $1\%$ of the data. 
    
Timing results for $\gamma=0.05$ are in Table~\ref{table:timing}. Feature extraction reduces dimension instead of increasing sparsity, and while both algorithms take roughly the same number of flops, feature extraction has roughly a $2\times$ edge in speed since it has simpler data structures which have better data locality and can be exploited by many algorithms.  The timing results are broken down into fine detail to show that the majority of time is in the actual K-means iteration on the reduced/sparsified data. We also note that without preconditioning, K-means never converges within 100 iterations in our tests, a sign that it does not capture the structure of the data, and this greatly contributes to the runtime of the algorithm.

\begin{table}
    \caption{From the same $n=6\cdot 10^5$ simulations as Fig.~\ref{fig:mnist8mError}, at $\gamma=0.05$. All numbers
        are averages over the 10 trials, times in seconds. The K-means algorithm was limited to 100 iterations, and an asterisk denotes the algorithm never converged within this limit (on all trials). \label{table:timing} }
        \centering
\begin{tabular}{lcccc}
\toprule
Algorithm & Total time & Time to sample & Time to precondition & Iterations of K-means \\
 \midrule
Sparsified K-means & 228.8 s & 6.0 s & 33.7 s & 42.1 \\ 
Sparsified K-means, 2 pass & 237.0 s & 6.0 s & 33.7 s & 42.1 \\
Sparsified K-means, no preconditioning & 665.6 s & 5.1 s & NA & 100.0* \\
Feature extraction & 123.1 s & 0.4 s & NA & 41.9 \\
\bottomrule
\end{tabular} 
\end{table}

\paragraph{Out-of-core memory with $n=10^7$}
We implement out-of-core versions of the sparsified K-means algorithm and the feature extraction algorithm, which efficiently load and compress the data in a batched manner such that the entire matrix is never loaded all at once into the main memory of the computer. The dataset is created using the same ``Infinite MNIST'' code, so the setup is the same as the previous sections, i.e., $p=784$, but now $n=9,\!631,\!605$, having $3,\!168,\!805$, $3,\!280,\!085$ and $3,\!182,\!715$ examples of the digits ``0'', ``3'' and ``9'', respectively. Stored in double-precision using Matlab's default compression, which automatically reduces precision if possible, the matrix is 4.9~GB, and would be 56~GB if loaded into memory in double-precision. 
 When loading, the matrix is split into 58 chunks, each one (except the last) with dimensions $784 \times 167183$ and approximately $1$~GB in size.
We repeat the experiment three times, using $\gamma \in \{0.01,0.05\}$, and $10$ replicates.

Table~\ref{table:timingBigger} shows the results. 
Accuracy is similar to the $n=600,\!000$ simulation, which is not surprising since the data for both experiments were algorithmically generated from the Infinite MNIST code. 
As expected, time to load data from disk was significant. For example, in the second pass over data at $\gamma=.05$, loading the data required $125$ seconds while the actual time for computing for the updated mean and assignment, i.e.~Alg.~\ref{algo:kmeans2} except line 2, was just $28$ seconds.
However, the time to load the data from disk is still not a bottleneck in the overall computation time, since we only need to perform this once (or twice). 
In a distributed data setting, where loading the data is even more costly but we would also be able to take $\gamma$ very small, one may prefer the 1-pass variant over the 2-pass variant. 

The time to sample is non-negligible since it requires many calls to a pseudo-random number generator and creating sparse arrays. Preconditioning also takes $170$ seconds using the DCT, though we remark that the DCT in Matlab is not well-implemented, and Matlab performs an FFT on the same data in under 1 minute since it directly calls the fftw library. Directly calling fftw's DCT routine would likely be much faster; it would also be possible to accelerate this step using general-purpose graphical processing units (GP-GPU) since applying the DCT to an array is efficiently parallelized. 

For a general idea of how long K-means would take without sub-sampling, we run a single iteration of K-means and compare the results in Table~\ref{table:timingBigger2}, finding that we reduce the computational time by a factor of almost $40$. The actual time for the full K-means algorithm would not improve by quite this much, since it is not clear if the number of iterations would change, and there is also the fixed cost of loading the data once (for sparsified K-means) or twice (for the 2-pass variant).
We also caution that this was a single run of a single iteration of K-means, so the factor $40$ is of limited precision, but overall, this is better than the result we hope to see since we have $\gamma=0.05$ meaning we have kept $1/20^\text{th}$ of the data. One reason for seeing $40$ times speedup instead of $20$ times speedup may be that Matlab paged memory onto secondary storage for the full version of the algorithm, or that the sparsified data was small enough to fit entirely inside the CPU cache memory (20 MB) rather than RAM.

\begin{table}
    \caption{From the $n=9,\!631,\!605$ simulation.
 All numbers are averages over the 3 trials, times in seconds. Accuracy listed as mean $\pm $ standard deviation.  \label{table:timingBigger} }
    \centering
    \begin{tabular}{clcccccc}
        \toprule
       & & & & \multicolumn{4}{c}{Time} \\
        \cmidrule(l){5-8}
        & Algorithm & Accuracy & Iterations of K-means& Total & to sample & to precondition & to load data from disk \\
        \midrule
\multirow{3}{*}{$\gamma=.01\;\Bigg\{$}&\multicolumn{1}{l}{Sparsified K-means} 		   & $0.745\pm .0008$ & 100* & 2630s & 38s & 170s & 125s  \\
&\multicolumn{1}{l}{Sparsified K-means, 2 pass} & $0.927\pm .0018$ & 100* & 2783s & 38s & 170s & 250s  \\
&\multicolumn{1}{l}{Feature extraction}		   & $0.680\pm .0610$ & 73.8 & 1123s & 18s & NA & 128s  \\
        \cmidrule(l){2-8}
\multirow{3}{*}{$\gamma=.05\;\Bigg\{$}&\multicolumn{1}{l}{Sparsified K-means} 		   & $0.887\pm .0002$ & 53.5 & 4380s & 137s & 159s & 126s  \\
&\multicolumn{1}{l}{Sparsified K-means, 2 pass} & $0.933\pm .0001$ & 53.5 & 4538s & 137s & 159s  & 254s  \\
&\multicolumn{1}{l}{Feature extraction}		   & $0.836\pm .0714$ & 70.4 & 3384s & 62s & NA & 129s  \\
        \bottomrule
    \end{tabular} 
\end{table}

 \begin{table}
     \caption{Estimated speedup. From the $n=9,\!631,\!605$ simulation, at $\gamma=0.05$.
          \label{table:timingBigger2} }
        \centering
        \begin{tabular}{lrrrrrr}
            \toprule
           & \multicolumn{2}{c}{Time to find assignments} & \multicolumn{2}{c}{Time to update all centers} & \multicolumn{2}{c}{Combined time} \\
          \cmidrule(rl){2-3}  \cmidrule(rl){4-5} \cmidrule(l){6-7} 
          Algorithm & Absolute & Speedup & Absolute & Speedup  & Absolute & Speedup\\
            \midrule
K-means & $130.0$s & $1\times$ & $150.8$s & $1\times$ & $280.8$s & $1\times$\\
Sparsified K-means & $1.3$s & $100\times$ & $5.7$s & $26.4\times$ &$7.0$s & $40.1\times$\\ 
            \bottomrule
        \end{tabular} 
    \end{table}

\subsection{Discussion}
On the MNIST data, all the fast algorithms show great speedup over standard K-means, and tunable accuracies that can reach the accuracy of standard K-means as $\gamma \rightarrow 1$. In our tests, the one-pass (preconditioned) sparsified K-means algorithm appears to be slightly more accurate than feature extraction, and significantly more accurate than feature selection and the non-preconditioned sparsified K-means.
In addition, our sparsified K-means has significantly lower variance than feature extraction, which means that the output of our method is more reliable and closer to the output of K-means on the full data among different initializations.

Furthermore, based on the MNIST experiments, if one can afford two passes over the data, the accuracy of our two-pass sparsified K-means reaches the accuracy of standard K-means and, at the same time, accurately estimates the cluster centers. For in-core problems, there is negligible cost for the extra pass, so the two-pass variant is the best candidate. For out-of-core problems, the one-pass variant may be preferred.

\section{Conclusions}\label{sec:conclusions}
We have presented a compression scheme for large-scale data sets which leads to both computational time and memory benefits in unsupervised learning tasks such as PCA and K-means clustering. A main feature of our approach is that it requires just one pass over the data thanks to the randomized preconditioning transformation, which makes it applicable to streaming and distributed data settings. In fact, the preconditioning transformation is an essential component of our approach which allows us to achieve accurate and reliable estimates in the data sparsification process and eliminates the need to revisit past entries of the data. A side-benefit of the preconditioning is a reduction in the variance of estimates.

Our sparsified K-means algorithm returns both assignments and cluster centers in a single pass over the data, whereas the state-of-the-art feature-based algorithms require at least two passes. Moreover, our approach leads to per-step guarantees on the clustering structure, as opposed to the guarantees on the overall objective function in feature-based algorithms. 

Finally, our compression scheme has the potential to be applicable in many other techniques in signal processing and machine learning, such as subspace learning, K-nearest neighbors, soft K-means, mixture models, and expectation-maximization algorithms. In these settings, the preconditioning and sampling technique could be used to either speed up computation for in-core memory problems, or to create one-pass variants for out-of-core or streaming problems.

\section*{Acknowledgment}
{\small
It is a pleasure to thank Michael Wakin and Alex Gittens for informative discussions, and the anonymous referees for constructive comments.}

\appendices

\setcounter{thm}{0}
\renewcommand{\thethm}{\Alph{section}\arabic{thm}}

\section{Exponential Concentration Inequalities}\label{sec:concentration_ineq}
Below are standard inequalities listed in expedient formats.

\begin{thm}[Hoeffding's Inequality \cite{ConcentrationInequalities}]\label{thm:Hoeffding}
    Let $z_{1},\ldots,z_{n}$ be independent, centered random variables, and assume that each one is bounded: 
    \[
    \mathbb{E}\left[z_{k}\right]=0\;\text{and}\;\left|z_{k}\right|\leq L_k\;\text{for each}\; k=1,\ldots,n.
    \]
    Introduce the sum $S=\sum_{k=1}^{n}z_{k}$, and let $\sigma^{2}=\sum_{k=1}^{n}L_k^2$. Then, for all $t\geq0$:
    \[
    \mathbb{P}\left\{ \left|S\right|\geq t\right\} \leq2\exp\left(\frac{-\nicefrac{t^{2}}{2}}{\sigma^{2}}\right).
    \]
\end{thm}

\begin{thm}[Bernstein Inequality \cite{ConcentrationInequalities}]\label{thm:Bernstein}
	Let $z_{1},\ldots,z_{n}$ be independent, centered random variables, and assume that each one is uniformly bounded: 
	\[
	\mathbb{E}\left[z_{k}\right]=0\;\text{and}\;\left|z_{k}\right|\leq L\;\text{for each}\; k=1,\ldots,n.
	\]
	Introduce the sum $S=\sum_{k=1}^{n}z_{k}$, and let $\sigma^{2}=\sum_{k=1}^{n}\mathbb{E}[z_{k}^{2}]$
	denote the variance of the sum. Then, for all $t\geq0$:
	\[
	\mathbb{P}\left\{ \left|S\right|\geq t\right\} \leq2\exp\left(\frac{-\nicefrac{t^{2}}{2}}{\sigma^{2}+\nicefrac{Lt}{3}}\right).
	\]
\end{thm}

\begin{thm}[Matrix Bernstein Inequality \cite{MatrixBernstein}]\label{thm:MatrixBernstein}
	Let $\mathbf{Z}_{1},\ldots,\mathbf{Z}_{n}$ be independent, symmetric,
	centered random matrices with dimension $p$, and assume that each
	one is uniformly bounded:
	\[
	\mathbb{E}\left[\mathbf{Z}_{k}\right]=\mathbf{0}\;\text{and}\;\left\Vert \mathbf{Z}_{k}\right\Vert _{2}\leq L\;\text{for each }k=1,\ldots,n.
	\]
	Introduce the sum $\mathbf{S}=\sum_{k=1}^{n}\mathbf{Z}_{k}$ , and
	let $\sigma^{2}=\left\Vert \sum_{k=1}^{n}\mathbb{E}\left[\mathbf{Z}_{k}^{2}\right]\right\Vert _{2}$ denote the variance. Then, for all $t\geq0$:
	\[
	\mathbb{P}\left\{ \left\Vert \mathbf{S}\right\Vert _{2}\geq t\right\} \leq p\exp\left(\frac{-\nicefrac{t^{2}}{2}}{\sigma^{2}+\nicefrac{Lt}{3}}\right).
	\]
\end{thm}

\section{Properties of the Sampling Matrix}\label{appendix:Properties-sampling}
\begin{thm}\label{thm:Properties_Sketching}
	Consider a sampling matrix $\mathbf{R}=[\mathbf{r}_{1},\ldots,\mathbf{r}_{m}]\in\mathbb{R}^{p\times m}$,
	where the $m$ columns are chosen uniformly at random from the set of all
	$p$ canonical basis vectors without replacement. Then, these columns form an orthonormal basis, i.e., $\mathbf{R}^{T}\mathbf{R}=\mathbf{I}_{m}$.
	Moreover, we have:
	\begin{equation}
	\mathbb{E}[\mathbf{RR}^{T}]=\frac{m}{p}\mathbf{I}_{p}\label{eq:expectation_RRT}
	\end{equation}
	and for any fixed vector $\mathbf{x}\in\mathbb{R}^{p}$ and $m\geq2$:
\begin{equation}
\mathbb{E}[\mathbf{RR}^{T}\mathbf{xx}^{T}\mathbf{RR}^{T}]\hspace{-1mm}=\hspace{-1mm}\frac{m(m-1)}{p(p-1)}\mathbf{xx}^{T}\hspace{-1mm}+\hspace{-1mm}\frac{m(p-m)}{p(p-1)}\diag(\mathbf{xx}^{T}).\label{eq:expectation_RRTxxTRRT}
\end{equation}
\end{thm}
\begin{IEEEproof}
	The columns of $\mathbf{R}$ are distinct canonical basis vectors, thus $\mathbf{R}^{T}\mathbf{R}=\mathbf{I}_{m}$. To prove~\eqref{eq:expectation_RRT}, note that $\mathbb{E}[\mathbf{RR}^{T}]=\sum_{i=1}^{m}\mathbb{E}[\mathbf{r}_{i}\mathbf{r}_{i}^{T}]$, and we will show that 
	\begin{equation}
	\mathbb{E}[\mathbf{r}_{i}\mathbf{r}_{i}^{T}]=\frac{1}{p}\mathbf{I}_{p}\;\text{for}\; i=1,\ldots,m.\label{eq:expectation_ri}
	\end{equation}
	The main difficulty is that the columns are dependent on each other since the sampling is without replacement. 
    Let us first consider $i=1$. In this case, the first column $\mathbf{r}_{1}$ is chosen uniformly at random from the set of all canonical basis vectors, i.e., $\mathbb{P}\left\{ \mathbf{r}_{1}=\mathbf{e}_{r_{1}}\right\}=\frac{1}{p}$ for $r_{1}=1,\ldots,p$. Thus 
	\[
	\mathbb{E}\left[\mathbf{r}_{1}\mathbf{r}_{1}^{T}\right]=\sum_{r_{1}=1}^{p}\mathbb{P}\left\{ \mathbf{r}_{1}=\mathbf{e}_{r_{1}}\right\} \mathbf{e}_{r_{1}}\mathbf{e}_{r_{1}}^{T}=\frac{1}{p}\sum_{r_{1}=1}^{p}\mathbf{e}_{r_{1}}\mathbf{e}_{r_{1}}^{T}=\frac{1}{p}\mathbf{I}_{p}.
	\]
	For $i \in \left\{2,\ldots,m\right\}$, we compute the expectation as follows:
	\begin{eqnarray*}
	\mathbb{E}\left[\mathbf{r}_{i}\mathbf{r}_{i}^{T}\right]	&  &\hspace{-5mm} =\mathbb{E}\left[\mathbb{E}\left[\mathbf{r}_{i}\mathbf{r}_{i}^{T}|\mathbf{r}_{1},\ldots,\mathbf{r}_{i-1}\right]\right]\\
  &  &\hspace{-5mm} =\sum_{\left(r_{1},\ldots,r_{i-1}\right)}\mathbb{P}\left\{ \mathbf{r}_{1}=\mathbf{e}_{r_{1}},\ldots\mathbf{r}_{i-1}=\mathbf{e}_{r_{i-1}}\right\} \\
		&  &\hspace{-5mm} \times \mathbb{E}\left[\mathbf{r}_{i}\mathbf{r}_{i}^{T}|\mathbf{r}_{1}=\mathbf{e}_{r_{1}},\ldots\mathbf{r}_{i-1}=\mathbf{e}_{r_{i-1}}\right]
	\end{eqnarray*}
	where the summation is over the set of $(i-1)$ distinct values from $\left\{1,\ldots,p\right\}$, thus $\mathbb{P}\{\mathbf{r}_{1}=\mathbf{e}_{r_{1}},\ldots,\mathbf{r}_{i-1}=\mathbf{e}_{r_{i-1}}\}=\frac{1}{p}\frac{1}{p-1}\ldots\frac{1}{p-(i-2)}$. Also, the expectation does not depend on the permutation of ${r_{1},\ldots,r_{i-1}}$, 
so we condense the sum to range over just the set, not permutation, of distinct values, and adjust by multiplying by  $\left(i-1\right)!$.
    Therefore,
	\begin{eqnarray*}
		\mathbb{E}\left[\mathbf{r}_{i}\mathbf{r}_{i}^{T}\right]&  &\hspace{-5mm}
		=\left(i-1\right)!\left(\frac{1}{p}\frac{1}{p-1}\ldots\frac{1}{p-\left(i-2\right)}\right) \\
        &  & \hspace{-5mm}
        \times \sum_{\substack{\{r_{1},\ldots,r_{i-1}\}\\\text{distinct}}}\Big(
        \sum_{\substack{r_{i}\;\text{from} \\ \text{remaining values}}}
        \mathbb{P}\left\{ \mathbf{r}_{i}=\mathbf{e}_{r_{i}}\right\} \mathbf{e}_{r_{i}}\mathbf{e}_{r_{i}}^{T}\Big)\\
		&  & \hspace{-5mm}\overset{(a)}{=}\left(i-1\right)!\left(\frac{1}{p}\frac{1}{p-1}\ldots\frac{1}{p-\left(i-2\right)}\frac{1}{p-\left(i-1\right)}\right) \\
  &  & \hspace{-5mm}
        \times \Big(\sum_{\substack{\{r_{1},\ldots,r_{i-1}\}\\\text{distinct}}}
        \sum_{\substack{r_{i}\;\text{from} \\ \text{remaining values}}}
        \mathbf{e}_{r_{i}}\mathbf{e}_{r_{i}}^{T}\Big)\\
		&  & \hspace{-5mm}\overset{(b)}{=}\left(i-1\right)!\left(\frac{1}{p}\frac{1}{p-1}\ldots\frac{1}{p-\left(i-2\right)}\frac{1}{p-\left(i-1\right)}\right)\\
		&  & \hspace{-5mm}
        \times \binom{p-1}{i-1}\mathbf{I}_{p}=\frac{1}{p}\mathbf{I}_{p}
	\end{eqnarray*}
	where (a) follows from $\mathbb{P}\{ \mathbf{r}_{i}=\mathbf{e}_{r_{i}}\}=\frac{1}{p-\left(i-1\right)}$ and (b) is obtained by counting the number of cases where $r_{i}=j$, $1\leq j\leq p$, and this can be easily computed by counting the number of cases that $j$ is not in the set $\{r_{1},\ldots,r_{i-1}\}$ which is $\binom{p-1}{i-1}$.
	This completes the proof of~\eqref{eq:expectation_RRT}. 
	
	Next, we show that \eqref{eq:expectation_RRTxxTRRT} holds. Note that:
	\begin{eqnarray*}
    &  & \hspace{-4mm}\mathbb{E}\left[\mathbf{RR}^{T}\mathbf{xx}^{T}\mathbf{RR}^{T}\right]\!\!=\!\!\!\sum_{\left(r_{1},\ldots,r_{m}\right)}\mathbb{P}\left\{ \mathbf{r}_{1}=\mathbf{e}_{r_{1}},\ldots\mathbf{r}_{m}=\mathbf{e}_{r_{m}}\right\} \\
		&  & 
        \times \Big(\sum_{i=1}^{m}\mathbf{e}_{r_{i}}\mathbf{e}_{r_{i}}^{T}\Big)\mathbf{xx}^{T}\Big(\sum_{i=1}^{m}\mathbf{e}_{r_{i}}\mathbf{e}_{r_{i}}^{T}\Big)\\
		&  & \hspace{-4mm}=\Big(\frac{1}{p}\frac{1}{p-1}\ldots\frac{1}{p-\left(m-1\right)}\Big)\left\{ \alpha_{1}\sum_{k=1}^{p}\mathbf{e}_{k}\mathbf{e}_{k}^{T}\mathbf{xx}^{T}\mathbf{e}_{k}\mathbf{e}_{k}^{T}
        \vphantom{+\alpha_{2}\sum_{k\neq l}\mathbf{e}_{k}\mathbf{e}_{k}^{T}\mathbf{xx}^{T}\mathbf{e}_{l}\mathbf{e}_{l}^{T}} 
        \right.\\
		&  & 
        \left.+\alpha_{2}\sum_{k\neq l}\mathbf{e}_{k}\mathbf{e}_{k}^{T}\mathbf{xx}^{T}\mathbf{e}_{l}\mathbf{e}_{l}^{T}\right\} 
	\end{eqnarray*}
	where the summation is over the set of $m$ distinct values from $\left\{1,\ldots,p\right\}$ and we should find the coefficients $\alpha_{1}$ and $\alpha_{2}$. In fact, $\alpha_{1}$ represents the number of cases that each $k$, $1\leq k \leq p$, is among the $m$ numbers chosen from $\{1,\ldots,p\}$ without replacement. Let's fix $r_{1}=1$, we then have $\binom{p-1}{m-1}(m-1)!$ cases. Thus, we see that:
	\[
	\alpha_{1}=\left[\binom{p-1}{m-1}\left(m-1\right)!\right]m=\frac{m\left(p-1\right)\left(p-2\right)!}{\left(p-m\right)!}.
	\]
	Similarly, $\alpha_{2}$ represents the number of cases that each pair of $k$ and $l$, $1\leq k,l \leq p$ and $k \neq l$, is among the $m$ numbers chosen from $\{1,\ldots,p\}$.  Let's fix $r_{1}=1$ and $r_{2}=2$, we then have $\binom{p-2}{m-2}(m-2)!$ cases. This argument leads to:
	\[
	\alpha_{2}\!=\!\left[\binom{p-2}{m-2}\left(m-2\right)!\right]m\left(m-1\right)\!=\!\frac{m\left(m-1\right)\left(p-2\right)!}{\left(p-m\right)!}.
	\]
	Since $\alpha_{2}<\alpha_{1}$, we can write the expectation as follows:
	\begin{eqnarray*}
		&  & \hspace{-5mm}\mathbb{E}\left[\mathbf{RR}^{T}\mathbf{xx}^{T}\mathbf{RR}^{T}\right]=\left(\frac{1}{p}\frac{1}{p-1}\ldots\frac{1}{p-\left(m-1\right)}\right)\\
		&  & \hspace{-5mm}\times\!\!\left(\!\alpha_{2}\Big(\sum_{k=1}^{p}\mathbf{e}_{k}\mathbf{e}_{k}^{T}\Big)\mathbf{xx}^{T}\Big(\sum_{k=1}^{p}\mathbf{e}_{k}\mathbf{e}_{k}^{T}\Big)\!+\!\left(\alpha_{1}-\alpha_{2}\right)\text{diag}\left(\mathbf{xx}^{T}\right)\!\right)
	\end{eqnarray*}
	and this completes the proof. 
\end{IEEEproof}
\begin{lem}\label{thm:prob-sample-without}
	Suppose we sample $m$ entries of $\x\in\R^p$ uniformly at random without replacement. Then, the probability of keeping the $j$-th entry of $\x$ is $\frac{m}{p}$ for all $j=1,\ldots,p$.
\end{lem}
\begin{IEEEproof}
	The proof follows from the properties of sampling matrices given in Theorem~\ref{thm:Properties_Sketching}. For a matrix $\RR\in\R^{p\times m}$ containing $m$ distinct canonical basis vectors, we show that $\E[\RR\RR^T]=\frac{m}{p}\mathbf{I}_p$.
	Each component of the sub-sampled data $\w=\RR\RR^T\x$ is a random variable with two possible values: the $j$-th entry of $\w$ takes the same value as the corresponding entry of $\x$ with probability $\pi_j$ and zero otherwise. Since $\E[\w]=\E[\RR\RR^T\x]=\frac{m}{p}\x$, we conclude that $\pi_j=\frac{m}{p}$ for $j=1,\ldots,p$.  
\end{IEEEproof}
\section{Proof of Theorem \ref{thm:covariance}}\label{sec:proof-of-covarinace}
We present the proof of Thm.~\ref{thm:covariance} on the covariance estimator. First, we show that $\Cn$ is an unbiased estimator, i.e., $\E[\Cn]=\Cemp$. Using Thm.~\ref{thm:Properties_Sketching}, we compute the following expectations:
\begin{equation}
\E[\Cemph]=\Cemp+\frac{(p-m)}{(m-1)}\diag(\Cemp)\label{eq:expectation-cemph}
\end{equation}
and 
\begin{equation}
\E[\diag(\Cemph)]\hspace{-1mm}=\hspace{-1mm}\diag(\E[\Cemph])\hspace{-1mm}=\hspace{-1mm}\frac{(p-1)}{(m-1)}\diag(\Cemp).\label{eq:expectation-cemph-diag}
\end{equation}
Hence, using~\eqref{eq:expectation-cemph} and \eqref{eq:expectation-cemph-diag}, we get $\E[\Cn]=\Cemp$. To find the closeness of $\Cn$ to its expectation $\Cemp$, we use the matrix Bernstein inequality (Thm.~\ref{thm:MatrixBernstein}). Note that $(\Cn-\Cemp)$ can be written as a sum of $n$ independent centered random matrices: 
\begin{equation}
\Cn-\Cemp=\sum_{i=1}^{n}\Z_i=\sum_{i=1}^{n}\frac{1}{n}(\Z_{i}^{(1)}-\Z_{i}^{(2)}-\Z_{i}^{(3)})
\end{equation}
where $\Z_{i}^{(1)}=\frac{p(p-1)}{m(m-1)}\w_i\w_i^T$, $\Z_{i}^{(2)}=\frac{p(p-m)}{m(m-1)}\diag(\w_i\w_i^T)$, $\Z_{i}^{(3)}=\x_i\x_i^T$, and $\w_i=\RR_i\RR_i^{T}\x_i$ is the sub-sampled data. To apply matrix Bernstein, we should find a uniform bound on the spectral norm of each summand $\|\Z_i\|_2$. We find a uniform bound for the spectral norm of $\Z_{i}^{(1)}$:
\begin{eqnarray}
\hspace{-10mm}&\|\Z_{i}^{(1)}\|_2\hspace{-3mm}&=\frac{p(p-1)}{m(m-1)}\|\w_i\w_i^{T}\|_2=\frac{p(p-1)}{m(m-1)}\|\w_i\|_2^{2} \nonumber\\
\hspace{-10mm}&\hspace{-3mm}&\leq \frac{p(p-1)}{m(m-1)} \rho \|\x_i\|_2^{2}\leq\frac{p(p-1)}{m(m-1)} \rho \|\X\|_\maxCol^2.
\end{eqnarray}
For the second term $\Z_{i}^{(2)}$, it is easy to verify that  $\diag(\w_i\w_i^{T})\preccurlyeq\diag(\x_i\x_i^{T})$, where $\mathbf{A}\preccurlyeq\mathbf{B}$ means that $\mathbf{B}-\mathbf{A}$ is positive semidefinite, and thus we get $\|\diag(\w_i\w_i^{T})\|_2\leq\|\diag(\x_i\x_i^{T})\|_2$ which can be used to bound $\|\Z_{i}^{(2)}\|_2$:
\begin{equation}
\|\Z_{i}^{(2)}\|_2\!=\!\frac{p(p-m)}{m(m-1)}\|\diag(\w_i\w_i^{T})\|_2\leq\frac{p(p-m)}{m(m-1)}\|\X\|_\text{max}^2.
\end{equation}
We also find a uniform bound for the spectral norm of $\Z_{i}^{(3)}$:
\begin{equation}
\|\Z_{i}^{(3)}\|_2=\|\x_i\x_i^{T}\|_2=\|\x_i\|_2^{2} \leq \|\X\|_\maxCol^2
\end{equation}
and the triangle inequality for the spectral norm leads to:
\begin{equation}
\|\Z_i\|_2\leq \frac{1}{n}\left(\|\Z_{i}^{(1)}\|_2+\|\Z_{i}^{(2)}\|_2+\|\Z_{i}^{(3)}\|_2\right)\leq L
\end{equation} 
where $L$ is given in~\eqref{eq:covariance-L}. 

Next, we compute the variance of our estimator $\sigma^2$:
\begin{equation}
\sigma^2=\|\E[(\Cn-\Cemp)^2]\|_2=\|\sum_{i=1}^{n}\E[\Z_i^2]\|_2.
\end{equation}
We find the variance $\sigma^2$ by first computing the expectation:
\begin{equation}
\E[\Z_i^2]=\frac{1}{n^2}\E[(\Z_{i}^{(1)}-\Z_{i}^{(2)}-\Z_{i}^{(3)})^2]
\end{equation}
which requires computing the expectation of the product of terms $\Z_{i}^{(1)}$, $\Z_{i}^{(2)}$, and $\Z_{i}^{(3)}$. We begin by computing:
\begin{eqnarray*}
&&\hspace{-8mm}\E[\Z_{i}^{(1)}\Z_{i}^{(1)}]\!\!=\!\! \frac{p^{2}(p-1)^2}{m^{2}(m-1)^2}\E[\w_i\w_i^{T}\w_i\w_i^{T}]\\
&&\hspace{-8mm}=\!\!\frac{p^{2}(p-1)^2}{m^{2}(m-1)^2}\E[\|\w_i\|_2^{2}\w_i\w_i^{T}]\!\preccurlyeq\!\! \frac{p^{2}(p-1)^2}{m^{2}(m-1)^2} \rho \|\x_i\|_2^{2}\E[\w_i\w_i^{T}]\\
&&\hspace{-8mm}=\!\!\frac{p(p-1)}{m(m-1)}\rho\|\x_i\|_2^{2}\x_i\x_i^{T}\!\!+\!\frac{p(p-1)(p-m)}{m(m-1)^2}\rho\|\x_i\|_2^{2}\diag(\x_i\x_i^{T})
\end{eqnarray*}
where the inequality follows the fact that expectation preserves the semidefinite order and we also used Thm.~\ref{thm:Properties_Sketching} to compute $\E[\w_i\w_i^{T}]$. Now, we compute $\E[\Z_{i}^{(2)}\Z_{i}^{(2)}]$:
\begin{eqnarray*}
&&\hspace{-8mm}\E[\Z_{i}^{(2)}\Z_{i}^{(2)}]\!=\!\frac{p^{2}(p-m)^2}{m^{2}(m-1)^2}\E[(\diag(\w_i\w_i^{T}))^2] \\
&&\hspace{-8mm}=\!\frac{p^{2}(p-m)^2}{m^{2}(m-1)^2}\frac{m}{p} (\diag(\x_i\x_i^{T}))^2\!=\!\frac{p(p-m)^2}{m(m-1)^2}(\diag(\x_i\x_i^{T}))^2
\end{eqnarray*}
where this follows from $\E[(\diag(\w_i\w_i^T))^2]=\frac{m}{p}(\diag(\x_i\x_i^T))^2$. To see this, let $w_{i,j}$ denote the $j$-th element of $\w_i$ and note that $\diag(\w_i\w_i^T)$ is a diagonal matrix where the $j$-th element is $w_{i,j}^2$. Thus, $(\diag(\w_i\w_i^T))^2$ is also a diagonal matrix where the $j$-th element is equal to $w_{i,j}^4$. Based on Lemma~\ref{thm:prob-sample-without}, the probability of keeping the $j$-th element of $\x_i$ under the uniform sampling without replacement is $\frac{m}{p}$, which means that $\E[w_{i,j}^4]=\frac{m}{p}x_{i,j}^4$.

Next, we can easily find the following expectations:   
\begin{eqnarray*}
&&\hspace{-8mm}\E[\Z_{i}^{(1)}\Z_{i}^{(3)}]=\E[\Z_{i}^{(3)}\Z_{i}^{(1)}]^{T}=\frac{p(p-1)}{m(m-1)}\E[\w_i\w_i^{T}]\x_i\x_i^{T}\\
&&\hspace{-8mm}=\|\x_i\|_2^{2}\x_i\x_i^{T}+\frac{(p-m)}{(m-1)}\diag(\x_i\x_i^{T})\x_i\x_i^{T}
\end{eqnarray*}
and 
\begin{eqnarray*}
	&&\hspace{-8mm}\E[\Z_{i}^{(2)}\Z_{i}^{(3)}]\!=\!\E[\Z_{i}^{(3)}\Z_{i}^{(2)}]^{T}\!=\!\frac{p(p-m)}{m(m-1)}\E[\diag(\w_i\w_i^{T})]\x_i\x_i^{T}\\
	&&\hspace{-8mm}=\!\frac{(p-m)}{(m-1)}\diag(\x_i\x_i^{T})\x_i\x_i^{T}
\end{eqnarray*}
and
\[
\E[\Z_{i}^{(3)}\Z_{i}^{(3)}]=\x_i\x_i^{T}\x_i\x_i^{T}=\|\x_i\|_2^{2}\x_i\x_i^{T}.
\]
Hence, based on the expectations computed above and the triangle inequality, we get:
\begin{eqnarray*}
	&&\hspace{-8mm}\sigma^2\!=\!\Big\|\sum_{i=1}^{n}\E[\Z_i^{2}]\Big\|_2\!\leq\! \frac{1}{n}\left(\frac{p(p-1)}{m(m-1)}\rho\!-1\!\!\right)\cdot\Big\|\frac{1}{n}\sum_{i=1}^{n}\|\x_i\|_2^{2}\x_i\x_i^{T} \Big\|_2 \\
	&&\hspace{-8mm}+\!\frac{1}{n}\frac{p(p-1)(p-m)}{m(m-1)^2}\rho\Big\|\frac{1}{n}\sum_{i=1}^{n}\|\x_i\|_2^{2}\diag(\x_i\x_i^{T}) \Big\|_2\\
	&&\hspace{-8mm}+\!\frac{1}{n^2}\frac{p(p-m)^2}{m(m-1)^2}\Big\|\sum_{i=1}^{n}(\diag(\x_i\x_i^{T}))^2\Big\|_2\\
	&&\hspace{-8mm}+\!\frac{1}{n^2}\left(\big\|\sum_{i=1}^{n}\E[\Z_i^{(1)}\Z_i^{(2)}]\big\|_2+ \big\|\sum_{i=1}^{n}\E[\Z_i^{(2)}\Z_i^{(1)}]\big\|_2\right).
\end{eqnarray*}
We also have the following two inequalities:
\[
\frac{1}{n}\sum_{i=1}^{n}\|\x_i\|_2^{2}\x_i\x_i^{T}\preccurlyeq\|\X\|_\maxCol^2\cdot\Cemp
\]
and 
\[
\frac{1}{n}\sum_{i=1}^{n}\|\x_i\|_2^{2}\diag(\x_i\x_i^{T})\preccurlyeq\|\X\|_\maxCol^2\cdot\diag(\Cemp).
\]
In the last step, we find an upper bound for the following:
\begin{eqnarray*}
&&\hspace{-8mm}\Big\|\sum_{i=1}^{n}\E[\Z_i^{(1)}\Z_i^{(2)}]\Big\|_2\leq\sum_{i=1}^{n}\|\E[\Z_i^{(1)}\Z_i^{(2)}]\|_2\\
&&\hspace{-8mm}\leq \sum_{i=1}^{n}\E[\|\Z_i^{(1)}\Z_i^{(2)}\|_2]\leq\sum_{i=1}^{n}\E[\|\Z_i^{(1)}\|_2\|\Z_i^{(2)}\|_2]
\end{eqnarray*}
where this follows from the triangle inequality, Jensen's inequality, and the fact that for two symmetric matrices $\mathbf{A}$ and $\mathbf{B}$, we have $\|\mathbf{A}\mathbf{B}\|_2\leq\|\mathbf{A}\|_2\|\mathbf{B}\|_2$. We compute the two terms inside the expectation: 
\[
\|\Z_i^{(1)}\|_2=\frac{p(p-1)}{m(m-1)}\|\w_i\|_2^{2}=\frac{p(p-1)}{m(m-1)}\x_i^{T}\RR_i\RR_i^{T}\x_i
\]
and 
\[
\|\Z_i^{(2)}\|_2=\frac{p(p-m)}{m(m-1)}\|\diag(\w_i\w_i^T)\|_2\leq\frac{p(p-m)}{m(m-1)}\|\X\|_\text{max}^{2}.
\]
Hence, using the property $\E[\RR_i\RR_i^{T}]=\frac{m}{p}\eye_p$, we get: 
\begin{eqnarray*}
&&\hspace{-8mm}\E[\|\Z_i^{(1)}\|_2\|\Z_i^{(2)}\|_2]\!\leq\! \frac{p^{2}(p-1)(p-m)}{m^{2}(m-1)^{2}}\|\X\|_\text{max}^{2}\E[\x_i^{T}\RR_i\RR_i^{T}\x_i]\\
&&\hspace{-8mm}=\frac{p(p-1)(p-m)}{m(m-1)^{2}}\|\X\|_\text{max}^{2}\|\x_i\|_2^{2}
\end{eqnarray*}
and using $\|\X\|_F^{2}=\sum_{i=1}^{n}\|\x_i\|_2^{2}$, we have:
\[
\Big\|\sum_{i=1}^{n}\E[\Z_i^{(1)}\Z_i^{(2)}]\Big\|_2\leq \frac{p(p-1)(p-m)}{m(m-1)^{2}}\|\X\|_\text{max}^{2}\|\X\|_F^{2}
\]
and this completes the proof. 
\section{Preservation of Pairwise Distances}
\label{sec:pairwiseDistance}
\newcommand{\VV}{\mathbf{V}}
\newcommand{\cc}{\mathbf{c}}
\begin{thm}\label{thm:ROS-distance-preserve}
Let $\x_1$ and $\x_2$ be two fixed vectors in $\R^{p}$. Consider the structured dimension reduction map consisting of the preconditioning transformation $\Hadamard\Diag$~\eqref{eq:ROS} and the sampling matrix $\RR\in\R^{p\times m}$, where the $m$ columns are chosen uniformly at random from the set of all canonical basis vectors without replacement. Then, with probability at least $1-3\beta^{-1}$,
\begin{equation}
0.40\left\|\x_1-\x_2\right\|_2\leq\left\|\sqrt{\frac{p}{m}}\RR^{T}\Hadamard\Diag(\x_1-\x_2)\right\|_2\leq 1.48\left\|\x_1-\x_2\right\|_2 
\end{equation}
given that $4\left[\sqrt{\beta}+\sqrt{8\log(\beta p)}\right]^{2}\log(\beta)\leq m \leq p$.
\end{thm}
\begin{IEEEproof}
This result is a straightforward consequence of Theorem 3.1 in~\cite{ImprovedAnalysis}. Let us denote $\x=\x_1-\x_2$ and represent it as $\x=\VV\cc$, where $\VV\in\R^{p\times \beta}$, $\beta<m$, is an orthonormal matrix and $\cc\in\R^{\beta}$ (e.g.~the first column of $\VV$ is $\x/\|\x\|_2$ and the remaining $(m-1)$ columns can be chosen via Gram-Schmidt). We then have the following deterministic lower and upper bounds for $\|\RR^T\Hadamard\Diag\x\|_2=\|\RR^T\Hadamard\Diag\VV\cc\|_2$:
\[
\sigma_\beta(\RR^T\Hadamard\Diag\VV) \|\cc\|_2\leq\|\RR^T\Hadamard\Diag\VV\cc\|_2\leq \sigma_1(\RR^T\Hadamard\Diag\VV) \|\cc\|_2
\] 
where $\sigma_1$ and $\sigma_\beta$ denote the largest and smallest singular values. Based on~\cite{ImprovedAnalysis}, for $m\geq 4[\sqrt{\beta}+\sqrt{8\log(\beta p)}]^{2}\log(\beta)$ and with probability at least $1-3\beta^{-1}$,
\[
0.40\sqrt{m/p}\leq \sigma_\beta(\RR^T\Hadamard\Diag\VV),\;\;
\sigma_1(\RR^T\Hadamard\Diag\VV)\leq1.48\sqrt{m/p}.
\]
Note that $\|\cc\|_2=\|\VV\cc\|_2=\|\x\|_2$ since $\VV$ is an orthonormal matrix and this completes the proof.
\end{IEEEproof}

\bibliographystyle{IEEEtran}
\bibliography{PHD_Farhad}

\end{document}